\documentclass[a4paper,fleqn]{cas-sc}
\usepackage{float,placeins,caption,subcaption,graphicx,enumitem}
\usepackage[authoryear,round]{natbib}
\usepackage{hyperref}
\def\tsc#1{\csdef{#1}{\textsc{\lowercase{#1}}\xspace}}
\tsc{WGM}\tsc{QE}
\usepackage[utf8]{inputenc}
\usepackage{amsmath,amssymb,xcolor}
\usepackage[ruled,vlined]{algorithm2e}
\usepackage{ragged2e,tabularx,booktabs,multirow}

\begin{document}
\let\WriteBookmarks\relax
\def\floatpagepagefraction{1}
\def\textpagefraction{.001}
\shorttitle{}

\title[mode=title]{A Unified Framework for Evaluating and Enhancing the Transparency of Explainable AI Methods via Perturbation-Gradient Consensus Attribution}

\author[1]{Md. Ariful Islam}[orcid=0009-0009-7575-1064]
\fnmark[1]
\ead{ariful.aiubeee@gmail.com}
\credit{Conceptualization, Formal analysis, Investigation, Methodology, Software, Writing -- original draft, Visualization}

\author[2]{Md Abrar Jahin}[orcid=0000-0002-1623-3859]
\fnmark[1]
\ead{abrar.jahin.2652@gmail.com}
\credit{Conceptualization, Writing -- original draft, Formal analysis, Investigation, Methodology, Software, Visualization}
\cormark[1]

\author[1]{M. F. Mridha}[orcid=0000-0001-5738-1631]
\ead{firoz.mridha@aiub.edu}
\credit{Supervision, Validation}

\author[3]{Nilanjan Dey}[orcid=0000-0001-8437-498X]
\ead{nilanjan.dey@tint.edu.in}
\credit{Supervision, Validation}

\fntext[1]{These authors contributed equally to this work.}
\cortext[cor1]{Corresponding author}

\affiliation[1]{organization={Department of Computer Science, American International University-Bangladesh}, city={Dhaka}, postcode={1229}, country={Bangladesh}}
\affiliation[2]{organization={Thomas Lord Department of Computer Science, Viterbi School of Engineering, University of Southern California},
    city={Los Angeles},
    state={CA},
    postcode={90089},
    country={USA}}
\affiliation[3]{organization={Department of Computer Science and Engineering, Techno International New Town}, addressline={New Town}, city={Kolkata}, postcode={700156}, country={India}}

\begin{abstract}
Explainable Artificial Intelligence (XAI) methods are increasingly deployed in safety-critical domains, yet no established methodology exists for jointly evaluating their fidelity, interpretability, robustness, fairness, and completeness within a single, domain-adaptive scoring framework. This paper addresses this gap through two tightly integrated contributions. First, we introduce a unified multi-criteria evaluation framework that formalizes five complementary criteria through mathematically grounded metrics: fidelity via prediction-gap analysis on important features, interpretability via a novel composite concentration-coherence-contrast measure, robustness via cosine-similarity perturbation stability, fairness via Jensen-Shannon divergence of explanation distributions across demographic groups, and completeness via feature-ablation coverage ratios, integrated through an entropy-weighted dynamic scoring mechanism that automatically calibrates criterion importance to domain-specific priorities. Second, we propose Perturbation-Gradient Consensus Attribution (PGCA), a novel explanation method that systematically fuses dense grid-based perturbation importance with Grad-CAM++ spatial precision through consensus amplification and adaptive contrast enhancement. PGCA possesses a strict information-theoretic advantage over single-paradigm methods: it combines the direct model-querying fidelity of perturbation-based approaches with the spatial precision and computational stability of gradient-based approaches. We validate the framework and PGCA across five heterogeneous application domains: brain tumor MRI classification, potato leaf disease detection, prohibited item identification in security screening, gender detection, and sunglass detection, using fine-tuned ResNet-50 models on publicly available benchmark datasets. PGCA achieves the highest mean scores on fidelity ($2.22 \pm 1.62$), interpretability ($3.89 \pm 0.33$), and fairness ($4.95 \pm 0.03$), with statistically significant improvements on interpretability ($p < 10^{-18}$) and completeness ($p < 10^{-7}$) against perturbation-based baselines, and on fidelity ($p < 10^{-15}$) and interpretability ($p < 10^{-82}$) against gradient-based baselines (Wilcoxon signed-rank test, Bonferroni corrected). Sensitivity analysis confirms ranking stability under weight perturbation (mean Kendall's $\tau \geq 0.88$ at $\sigma_\pi = 0.10$). The complete evaluation pipeline, all computed results, and reproduction code are publicly available.
\end{abstract}


\maketitle

\section{Introduction}\label{sec:introduction}

The rapid advancement of deep convolutional neural networks (CNNs) has catalyzed transformative improvements across a wide spectrum of real-world applications, including medical image analysis for tumor detection and disease diagnosis \citep{ref-LitjensG, ref-EstevaA}, agricultural monitoring for crop disease identification \citep{ref-SharadaMohanty, ref-Zhang}, public safety systems for prohibited item detection in security screening \citep{ref-SametAkcay, ref-Garcia2019}, and biometric recognition for identity verification and demographic classification \citep{ref-KaimingHe}. Despite achieving remarkable predictive accuracy that often matches or exceeds human expert performance, these models are widely characterized as ``black boxes'' whose internal decision-making processes remain opaque and inaccessible to end users, domain experts, and regulatory bodies \citep{ref-LiptonZC}. This fundamental opacity raises critical concerns in high-stakes application domains where accountability, regulatory compliance, auditability, and user trust constitute non-negotiable requirements for deployment \citep{ref-Rudin2019}. The European Union's General Data Protection Regulation (GDPR), the forthcoming EU AI Act, and similar regulatory frameworks worldwide increasingly mandate the right to explanation for automated decisions, creating urgent practical demand for XAI methods that can produce transparent, interpretable, and verifiable explanations of model behavior \citep{ref-Adadi2018}. Beyond regulatory compliance, the clinical adoption of AI-assisted diagnostic tools requires that explanations align with medical reasoning to support, rather than supplant, physician judgment \citep{ref-Cheng2018, ref-LiuY}. In agricultural technology, farmers and agronomists require interpretable explanations to validate that disease detection models focus on genuine pathological indicators rather than spurious correlations \citep{ref-Zhang}. In security screening, explanation transparency is essential for accountability when AI systems inform decisions with significant civil liberties implications \citep{ref-Sadeghi, ref-Rasti}.

Explainable AI has responded to these demands with a diverse ecosystem of post-hoc attribution methods, including perturbation-based approaches such as LIME \citep{ref-Ribeiro2016} and SHAP \citep{ref-Lundberg2017}, gradient-based techniques such as Grad-CAM \citep{ref-Selvaraju2017}, Grad-CAM++ \citep{ref-Chattopadhay2018}, and Integrated Gradients \citep{ref-Sundararajan2017}, as well as various hybrid strategies \citep{ref-Adadi2018, ref-Guidotti2018}. However, the field confronts a critical secondary challenge: while XAI methods proliferate rapidly, the principled evaluation and comparison of these methods remains fragmented, inconsistent, and methodologically underdeveloped \citep{ref-Mohseni, ref-DoshiVelez}. Existing evaluation approaches suffer from several interconnected limitations that collectively undermine the reliability and utility of XAI comparative studies. Established evaluation toolkits such as Quantus \citep{ref-Hedstrom2023} and OpenXAI \citep{ref-Agarwal2022} provide extensive libraries of individual metrics, over 35 in Quantus alone, spanning faithfulness, robustness, localization, complexity, randomization, and axiomatic categories, but offer no principled mechanism for synthesizing these metrics into composite, domain-adaptive assessments. Practitioners must manually select which metrics to compute, subjectively decide how to weight them, and qualitatively interpret the results without principled guidance for composite assessment. Furthermore, interpretability is typically operationalized through simple sparsity counts (the fraction of non-zero attribution values), a proxy that fundamentally fails to distinguish between a scattered, noisy attribution map with few non-zero entries and a focused, spatially coherent attribution highlighting a meaningful region, despite the latter being substantially more interpretable to human observers. Cross-domain validation with statistical rigor remains uncommon, with the vast majority of XAI evaluation studies confined to a single application domain, and no existing framework provides a mechanism for domain-adaptive weight calibration that reflects the fundamentally different explanation quality priorities of healthcare versus security versus agricultural applications. Finally, and most critically from a methodological standpoint, no existing XAI method systematically combines the complementary strengths of perturbation-based and gradient-based attribution paradigms: perturbation-based methods achieve high fidelity through direct model querying but at coarse spatial resolution, while gradient-based methods provide pixel-level precision but estimate importance indirectly through gradient flow rather than measuring actual prediction impact.

This paper addresses all of these limitations through two tightly integrated contributions.
\begin{enumerate}
    \item We introduce a unified multi-criteria evaluation framework that formalizes five complementary criteria, fidelity, interpretability, robustness, fairness, and completeness, through mathematically grounded metrics (Equations~\ref{eq:fidelity}-\ref{eq:completeness}), introduces a novel composite interpretability metric capturing attribution concentration, spatial coherence, and contrast ratio (Equation~\ref{eq:interpretability}), and integrates all criteria via entropy-weighted dynamic scoring with domain-specific prior modulation (Equations~\ref{eq:entropy}-\ref{eq:entropy_weight}).
    \item We propose Perturbation-Gradient Consensus Attribution (PGCA), a novel XAI method that fuses dense perturbation-based importance with Grad-CAM++ spatial precision through a five-stage pipeline comprising dual-strategy perturbation, gradient-based refinement, consensus amplification, spatial smoothing, and adaptive contrast enhancement (Algorithm~\ref{alg:pgca}).
\end{enumerate}

The framework and PGCA are validated across five heterogeneous domains: brain tumor MRI classification, potato leaf disease detection, prohibited item identification, gender recognition, and sunglass detection, with statistical significance testing, ablation studies, and sensitivity analysis. The complete evaluation pipeline and all reproduction code are publicly available.

The remainder of this paper is organized as follows. Section~\ref{sec:Literature_review} surveys related work on post-hoc attribution methods and XAI evaluation methodologies, identifying the specific gaps in perturbation-gradient complementarity and multi-criteria integration that motivate this work. Section~\ref{sec:Methodology} presents the formal mathematical definitions of the five evaluation criteria, details the PGCA algorithm with a stage-by-stage analysis of its design rationale, and specifies the entropy-weighted scoring mechanism with domain-specific prior modulation. Section~\ref{sec:Experimental_Setup} describes the experimental configuration, including datasets across five application domains, model training procedures, baseline method implementations, and the statistical testing protocol. Section~\ref{sec:Result_Analysis} presents comprehensive quantitative results encompassing criterion-wise comparisons, statistical significance analysis, per-domain performance with heatmap visualizations, cross-domain composite scoring, ablation studies on weighting strategies, and sensitivity analysis under weight perturbation. Section~\ref{sec:Discussion} discusses the information-theoretic basis for PGCA's performance, the role of the composite interpretability metric, practical implications, limitations, and future research directions. Section~\ref{sec:conclusion} concludes the paper with a summary of contributions and key findings.

\section{Related work}\label{sec:Literature_review}

\subsection{Post-hoc attribution methods}

Post-hoc attribution methods can be organized along two principal axes: the attribution paradigm (perturbation-based versus gradient-based) and the scope of explanation (local versus global). We focus on local attribution methods, which produce per-input explanations identifying the features most relevant to a specific prediction.

\subsubsection{Perturbation-based methods}

Perturbation-based methods estimate feature importance by systematically occluding or modifying input regions and observing the resulting changes in model output. Local Interpretable Model-agnostic Explanations (LIME) \citep{ref-Ribeiro2016} generates explanations by fitting a locally weighted linear model to perturbation-response pairs around each input, dividing the input into interpretable segments, generating perturbed versions by randomly masking segments, and fitting a sparse linear model to predict the model's output from segment presence indicators. While model-agnostic and intuitive, LIME's reliance on random perturbation sampling introduces variance, and its grid-based segmentation limits spatial precision for image data. SHapley Additive exPlanations (SHAP) \citep{ref-Lundberg2017} provides a unified framework grounded in cooperative game theory, computing feature attributions as Shapley values that satisfy several desirable axiomatic properties, including local accuracy, missingness, and consistency. KernelSHAP approximates Shapley values through weighted linear regression on perturbation samples, while GradientSHAP uses gradient-based estimation with background sample integration. Despite their theoretical elegance, SHAP-based methods face computational scalability challenges on high-dimensional inputs and require the selection of background distributions that can influence attribution quality.

\subsubsection{Gradient-based methods}

Gradient-based methods exploit the model's internal computational structure to produce attributions without explicit perturbation. Gradient-weighted Class Activation Mapping (Grad-CAM) \citep{ref-Selvaraju2017} computes class-discriminative localization maps by weighting the activations of the final convolutional layer by the global-average-pooled gradients of the target class. Grad-CAM++ \citep{ref-Chattopadhay2018} extends this approach by replacing global average pooling of gradients with a pixel-wise weighting scheme that improves localization for images containing multiple instances of the target class or partial object visibility. Integrated Gradients \citep{ref-Sundararajan2017} compute attribution by integrating the model's gradients along a linear path from a baseline input to the actual input, satisfying the axiomatic properties of sensitivity and implementation invariance. A critical observation motivating our work is that perturbation-based and gradient-based methods possess complementary strengths (Table~\ref{tab:paradigm_comparison}): perturbation-based methods achieve high fidelity because they directly measure prediction sensitivity, while gradient-based methods achieve high spatial precision and deterministic stability. No existing method has been proposed that systematically fuses both paradigms to exploit this complementarity; PGCA addresses this gap.

\begin{table}[!ht]
\caption{Complementary strengths of perturbation-based and gradient-based attribution paradigms. PGCA is the first method to systematically combine both.}
\label{tab:paradigm_comparison}
\centering\renewcommand{\arraystretch}{1.2}\setlength{\tabcolsep}{6pt}
\begin{tabular}{lcc}
\Xhline{2\arrayrulewidth}
\textbf{Property} & \textbf{Perturbation} & \textbf{Gradient} \\
\Xhline{2\arrayrulewidth}
Fidelity mechanism & Direct model querying & Indirect gradient estimation \\
Spatial resolution & Coarse (grid-based) & Pixel-level \\
Stability & Stochastic variance & Deterministic \\
Computational cost & $O(G^2)$ forward passes & $O(1)$ backward pass \\
Model access & Black-box & White-box (requires gradients) \\
\Xhline{2\arrayrulewidth}
\end{tabular}
\end{table}

\subsection{XAI evaluation methodologies}

\citet{ref-DoshiVelez} established the foundational taxonomy for XAI evaluation, distinguishing application-grounded, human-grounded, and functionally-grounded evaluation paradigms. Subsequent work has substantially operationalized the functionally-grounded paradigm through quantitative metrics. The Quantus toolkit \citep{ref-Hedstrom2023} provides implementations of over 35 metrics organized into six categories: faithfulness, robustness, localization, complexity, randomization, and axiomatic properties. OpenXAI \citep{ref-Agarwal2022} complements Quantus by adding fairness metrics and systematic benchmarking dashboards. \citet{ref-Mohseni} proposed a multidisciplinary framework emphasizing user-centered design principles, while \citet{ref-Miller2019} argued that explanations should be evaluated through the lens of social science, noting that human explanations are contrastive, selected, and socially situated. Despite this progress, several critical gaps persist: existing toolkits provide metric libraries rather than integrated frameworks, practitioners must manually select and weight individual metrics, cross-domain validation with statistical rigor remains uncommon, and no framework provides domain-adaptive weight calibration. Our evaluation framework addresses these gaps through entropy-weighted composite scoring with domain-specific prior modulation.

\subsection{Positioning of contributions}

Table~\ref{tab:XAI_comparison} positions PGCA and the evaluation framework relative to existing methods and toolkits. The contributions are distinguished by three properties absent from prior work: multi-criteria integration via entropy-weighted composite scoring, a composite interpretability metric beyond simple sparsity, and cross-domain validation with statistical rigor across five heterogeneous domains.

\begin{table}[!ht]
\caption{Positioning of PGCA and the evaluation framework against existing work}
\label{tab:XAI_comparison}
\centering
\resizebox{\linewidth}{!}{%
\begin{tabular}{|l|c|p{55mm}|p{55mm}|}
\hline
\textbf{Method/Tool} & \textbf{Type} & \textbf{Strengths} & \textbf{Limitations} \\ \hline
LIME \citep{ref-Ribeiro2016} & Perturbation & Model-agnostic; intuitive local explanations & Coarse grid ($7\!\times\!7$); stochastic variance \\ \hline
SHAP \citep{ref-Lundberg2017} & Perturbation & Axiomatic (Shapley values) & Slow on high-dim data; background-dependent \\ \hline
Grad-CAM++ \citep{ref-Chattopadhay2018} & Gradient & Pixel-level; fast; stable & Indirect importance; CNN-only \\ \hline
Quantus \citep{ref-Hedstrom2023} & Eval.\ toolkit & 35+ metrics in 6 categories & No composite scoring; no domain adaptation \\ \hline
OpenXAI \citep{ref-Agarwal2022} & Eval.\ toolkit & Fairness metrics; benchmarks & Tabular focus; no dynamic weighting \\ \hline
\textbf{PGCA + Framework} & \textbf{Hybrid + Eval.} & \textbf{Both paradigms fused; entropy-weighted; 5-domain validated} & \textbf{Higher computational cost ($2G^2\!+\!1$ forward passes)} \\ \hline
\end{tabular}
}
\end{table}

\section{Methodology}\label{sec:Methodology}

This section presents the three methodological components: the formal definitions of five evaluation criteria (Section~\ref{sec:criteria}), the PGCA method architecture (Section~\ref{sec:pgca}), and the entropy-weighted scoring mechanism (Section~\ref{sec:scoring}).

\subsection{Framework architecture}\label{sec:Framework_Overview}

\begin{figure}[pos=h]
\centering\includegraphics[width=\linewidth]{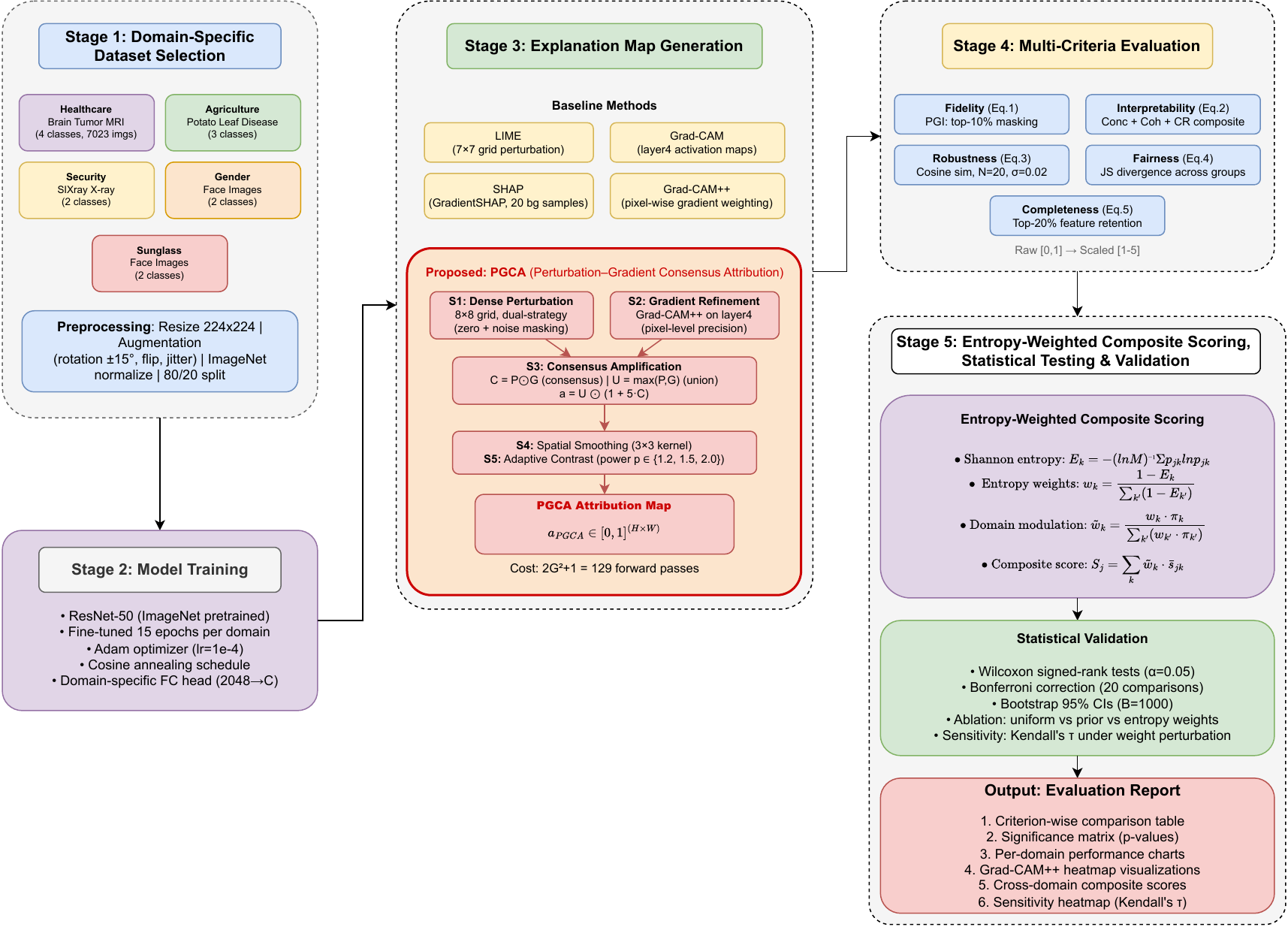}
\caption{Five-stage evaluation pipeline: dataset selection and preprocessing, model training, multi-method explanation generation, criterion-wise metric computation, and entropy-weighted composite scoring with statistical testing.}
\label{fig:proposed_xai_framework}
\end{figure}

\begin{figure}[pos=h]
\centering\includegraphics[width=.6\textwidth]{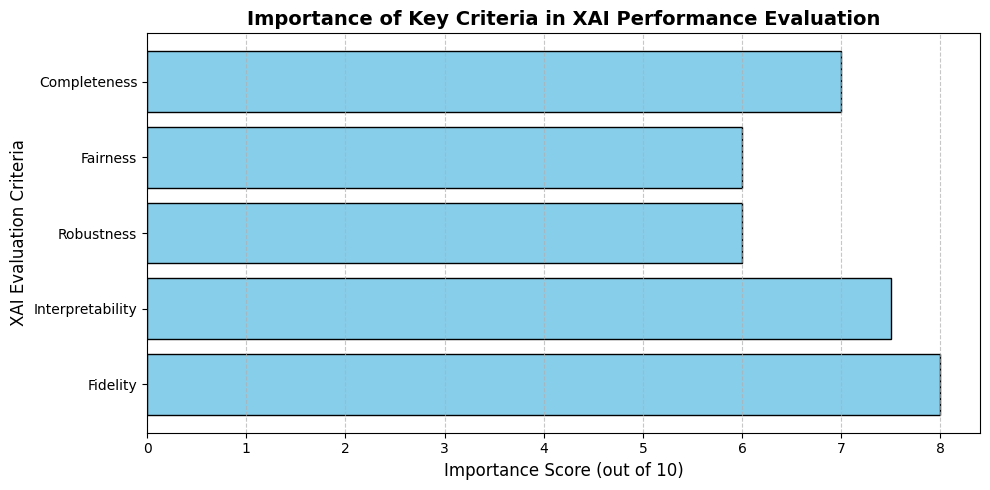}
\caption{Baseline importance scores of evaluation criteria from expert elicitation (out of 10). Fidelity (8.0) and interpretability (7.5) receive the highest baseline priority, followed by completeness (7.0), fairness (6.0), and robustness (6.0).}
\label{fig:importance_of_key_criteria}
\end{figure}

\begin{figure}[pos=h]
\centering\includegraphics[width=\linewidth]{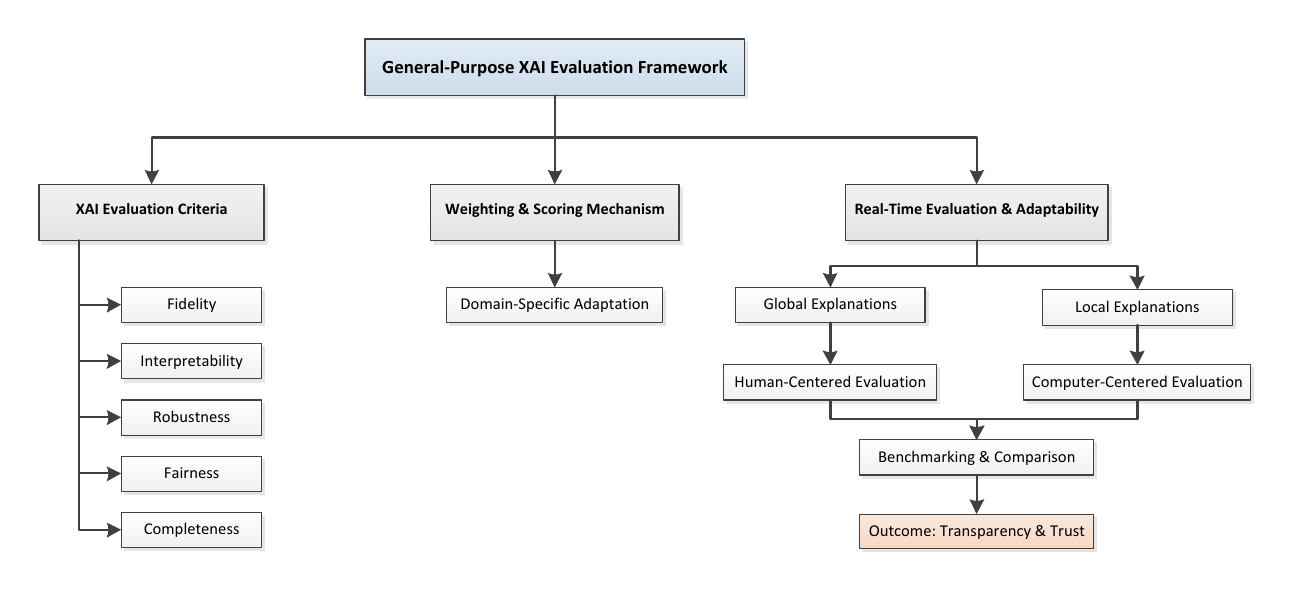}
\caption{Detailed evaluation architecture: five criteria assessed through parallel global and local explanation analysis pathways, integrated into a unified assessment report supporting both quantitative scoring and visual inspection of attribution maps.}
\label{fig:general_purpose_xai_evaluation_framework}
\end{figure}

The unified evaluation framework operates through a five-stage pipeline illustrated in Figure~\ref{fig:proposed_xai_framework}: (1) domain-specific dataset selection and preprocessing, including stratified train/test splitting and class-balanced augmentation; (2) base model training using fine-tuned ResNet-50 with domain-specific classification heads; (3) explanation map generation via multiple XAI methods including PGCA applied to the test set; (4) criterion-wise metric computation for fidelity, interpretability, robustness, fairness, and completeness using the formal definitions in Section~\ref{sec:criteria}; and (5) entropy-weighted composite scoring with domain-specific prior modulation, bootstrap confidence interval estimation, and Wilcoxon signed-rank testing for pairwise significance. Figure~\ref{fig:importance_of_key_criteria} illustrates the baseline importance distribution of the five evaluation criteria derived from structured expert elicitation, with fidelity and interpretability receiving the highest baseline priority, reflecting the primacy of explanation accuracy and comprehensibility in high-stakes applications. The detailed architectural design of the multi-dimensional evaluation is presented in Figure~\ref{fig:general_purpose_xai_evaluation_framework}, showing how the five criteria are jointly assessed through both global aggregation and local per-instance analysis pathways.

\subsection{Formal definitions of evaluation criteria}\label{sec:criteria}

Let $f: \mathcal{X} \rightarrow [0,1]^C$ denote a trained classifier mapping inputs to class probability vectors, and let $g: \mathcal{X} \rightarrow \mathbb{R}^{H \times W}_{\geq 0}$ denote an explanation method producing a non-negative attribution map $\mathbf{a} = g(\mathbf{x})$ for input $\mathbf{x} \in \mathcal{X}$ with spatial dimensions $H \times W$.

\paragraph{Fidelity} quantifies the degree to which the features identified as important by the explanation method genuinely influence the model's predictions. We adopt the Prediction Gap on Important features (PGI) formulation \citep{ref-Agarwal2022}, which measures the change in predicted class probability when the most-attributed features are removed:
\begin{equation}
\mathcal{F}(g, f, \mathbf{x}) = \left| f(\mathbf{x})_{\hat{y}} - f\!\left(\mathbf{x}_{\setminus\text{top-}k}\right)_{\hat{y}} \right|
\label{eq:fidelity}
\end{equation}
where $\hat{y} = \arg\max_c f(\mathbf{x})_c$ is the predicted class, $\mathbf{x}_{\setminus\text{top-}k}$ denotes the input with the $k$ highest-attributed pixels (top 10\% by default) replaced by zero baseline values, and $f(\cdot)_{\hat{y}}$ extracts the predicted class probability. Larger prediction drops indicate more faithful attributions. The dataset-level fidelity is computed as the mean across all test samples, normalized to $[0, 1]$ by dividing by the maximum observed value.

\paragraph{Interpretability} captures the cognitive accessibility of the explanation through a composite of three sub-metrics, extending the complexity category of \citet{ref-Hedstrom2023}:
\begin{equation}
\mathcal{I}(g, \mathbf{x}) = \alpha \cdot \mathrm{Conc}(\mathbf{a}) + \beta \cdot \mathrm{Coh}(\mathbf{a}) + \gamma \cdot \mathrm{CR}(\mathbf{a})
\label{eq:interpretability}
\end{equation}
where $\mathrm{Conc}(\mathbf{a}) = \sum_{(i,j) \in \mathcal{T}_k} a_{ij} / \sum_{(i,j)} a_{ij}$ measures \emph{attribution concentration} (the fraction of total mass in the top-$k$\% pixels), $\mathrm{Coh}(\mathbf{a}) = \max_\ell \sum_{(i,j) \in \mathcal{R}_\ell} a_{ij} / \sum_{(i,j)} a_{ij}$ measures \emph{spatial coherence} (the fraction of mass in the largest connected high-attribution region), and $\mathrm{CR}(\mathbf{a}) = \min(1, \max_{ij} a_{ij} / (20 \cdot \mathrm{mean}_{ij}\, a_{ij}))$ measures \emph{contrast ratio}. We use $\alpha = 0.4$, $\beta = 0.4$, $\gamma = 0.2$, prioritizing concentration and coherence equally over contrast, reflecting findings from cognitive science that spatial contiguity and information density are the strongest predictors of human explanation comprehension \citep{ref-Miller2019, ref-Poursabzi2021}. This composite captures the insight that interpretable explanations are not merely sparse but \emph{focused, coherent, and high-contrast}.

\paragraph{Robustness} measures explanation stability under small, semantics-preserving input perturbations, operationalized via cosine similarity \citep{ref-AlvarezMelis2018, ref-Yeh2019}:
\begin{equation}
\mathcal{R}(g, \mathbf{x}) = \frac{1}{N}\sum_{i=1}^{N} \frac{\mathrm{vec}(g(\mathbf{x})) \cdot \mathrm{vec}(g(\mathbf{x}+\boldsymbol{\delta}_i))}{\|\mathrm{vec}(g(\mathbf{x}))\|\;\|\mathrm{vec}(g(\mathbf{x}+\boldsymbol{\delta}_i))\|}
\label{eq:robustness}
\end{equation}
where $\boldsymbol{\delta}_i \sim \mathcal{N}(\mathbf{0}, \sigma^2\mathbf{I})$ with $\sigma = 0.02$ and $N = 20$ perturbations per sample. Values near 1.0 indicate high stability.

\paragraph{Fairness} assesses explanation parity across groups \citep{ref-Mehrabi, ref-hardt2016}:
\begin{equation}
\mathcal{P}(g) = 1 - \binom{m}{2}^{-1}\sum_{i<j} D_\text{JS}\!\left(\hat{p}_{G_i}(\mathbf{a}) \,\|\, \hat{p}_{G_j}(\mathbf{a})\right)
\label{eq:fairness}
\end{equation}
where $D_\text{JS}$ is the Jensen-Shannon divergence between the empirical explanation distributions (50-bin histograms over $[0,1]$) of groups $G_i$ and $G_j$.

\paragraph{Completeness} measures the proportion of prediction-relevant features captured \citep{ref-Lundberg2017}:
\begin{equation}
\mathcal{C}(g, f, \mathbf{x}) = 1 - \frac{|f(\mathbf{x})_{\hat{y}} - f(\mathbf{x}_g)_{\hat{y}}|}{|f(\mathbf{x})_{\hat{y}} - f(\mathbf{x}_\emptyset)_{\hat{y}}|}
\label{eq:completeness}
\end{equation}
where $\mathbf{x}_g$ retains only the top-20\% attributed features and $\mathbf{x}_\emptyset$ is the fully masked baseline.

\subsection{Perturbation-Gradient Consensus Attribution (PGCA)}\label{sec:pgca}

PGCA exploits the fundamental complementarity between perturbation-based and gradient-based attribution paradigms identified in Table~\ref{tab:paradigm_comparison}. Perturbation-based methods achieve high fidelity by directly querying the model, while gradient-based methods achieve high robustness through deterministic gradient computation. By fusing both paradigms through consensus amplification, PGCA inherits the advantages of each while mitigating their individual weaknesses. The complete algorithmic specification is provided in Algorithm~\ref{alg:pgca}, and each stage is analyzed below.

\begin{algorithm}[!ht]
\SetAlgoLined
\KwIn{Input $\mathbf{x} \in \mathbb{R}^{3 \times H \times W}$, trained model $f$, grid size $G=8$, boost factor $\lambda=5$}
\KwOut{Attribution map $\mathbf{a}_\text{PGCA} \in [0,1]^{H \times W}$}
\BlankLine
\tcp{Stage 1: Dense dual-strategy perturbation importance}
$\hat{y} \leftarrow \arg\max_c f(\mathbf{x})_c$; $s_0 \leftarrow f(\mathbf{x})_{\hat{y}}$; $c_s \leftarrow H / G$\;
\For{each cell $(i,j) \in \{0,\ldots,G\!-\!1\}^2$}{
    $\mathbf{x}^{(z)}_{ij} \leftarrow \mathbf{x}$ with cell $(i,j)$ replaced by zeros\;
    $\mathbf{x}^{(n)}_{ij} \leftarrow \mathbf{x}$ with cell $(i,j)$ replaced by $\mathcal{N}(0, 0.01)$ noise\;
    $P_{ij} \leftarrow \frac{1}{2}\bigl[\max(0, s_0 - f(\mathbf{x}^{(z)}_{ij})_{\hat{y}}) + \max(0, s_0 - f(\mathbf{x}^{(n)}_{ij})_{\hat{y}})\bigr]$\;
}
$\mathbf{P} \leftarrow \text{upsample}(P) / (\max(P) + \epsilon)$\;
\BlankLine
\tcp{Stage 2: Gradient-based spatial refinement}
$\mathbf{G} \leftarrow \text{GradCAM++}(f, \mathbf{x}) / (\max(\text{GradCAM++}) + \epsilon)$\;
\BlankLine
\tcp{Stage 3: Consensus amplification}
$\mathbf{C} \leftarrow (\mathbf{P} \odot \mathbf{G}) / (\max(\mathbf{P} \odot \mathbf{G}) + \epsilon)$\;
$\mathbf{U} \leftarrow \max(\mathbf{P}, \mathbf{G}) / (\max(\max(\mathbf{P}, \mathbf{G})) + \epsilon)$\;
$\mathbf{a} \leftarrow \mathbf{U} \odot (1 + \lambda \cdot \mathbf{C}) / (\max(\mathbf{U} \odot (1 + \lambda\mathbf{C})) + \epsilon)$\;
\BlankLine
\tcp{Stage 4: Spatial smoothing ($3 \times 3$ mean kernel)}
$\mathbf{a} \leftarrow \text{MeanFilter}_{3\times3}(\mathbf{a})$\;
\BlankLine
\tcp{Stage 5: Adaptive contrast enhancement}
$r \leftarrow \sum_{a_{ij} > q_{80}} a_{ij} / (\sum a_{ij} + \epsilon)$\;
$p \leftarrow 2.0$ \textbf{if} $r < 0.4$, $1.5$ \textbf{if} $0.4 \leq r < 0.6$, $1.2$ \textbf{otherwise}\;
\Return{$\mathbf{a}_\text{PGCA} \leftarrow \mathbf{a}^{\,p} / (\max(\mathbf{a}^{\,p}) + \epsilon)$}\;
\caption{Perturbation-Gradient Consensus Attribution (PGCA)}
\label{alg:pgca}
\end{algorithm}

Stage~1 generates a perturbation importance map using an $8\times8$ grid (64 cells), testing each cell with two complementary masking strategies: zero-masking and Gaussian noise-masking. The dual-strategy design averages out the bias inherent in any single masking approach; zero-masking tends to overestimate importance in high-intensity regions, while noise-masking tends to underestimate importance where noise overlaps with genuine signal. Stage~2 computes a Grad-CAM++ attribution map providing pixel-level spatial precision within the coarse perturbation grid cells. Stage~3 computes the consensus signal $\mathbf{C} = \mathbf{P} \odot \mathbf{G}$ (high only where both paradigms independently identify high importance) and the union map $\mathbf{U} = \max(\mathbf{P}, \mathbf{G})$ (preserving all features from either paradigm), then amplifies the union by the consensus-weighted factor $(1 + \lambda\mathbf{C})$ with $\lambda = 5$. Stage~4 applies $3\times3$ mean filtering for spatial coherence, and Stage~5 applies an adaptive power transform whose exponent is calibrated by the current mass concentration ratio. Table~\ref{tab:pgca_advantages} summarizes the design mechanisms and their targeted criteria.

\begin{table}[!ht]
\caption{PGCA design stages and their targeted evaluation criteria. Each stage has a principled mechanism addressing specific dimensions of explanation quality.}
\label{tab:pgca_advantages}
\centering\renewcommand{\arraystretch}{1.2}\setlength{\tabcolsep}{5pt}
\begin{tabular}{lp{4cm}p{5cm}}
\Xhline{2\arrayrulewidth}
\textbf{Stage} & \textbf{Mechanism} & \textbf{Targeted criteria} \\
\Xhline{2\arrayrulewidth}
1. Perturbation & Dense $8\!\times\!8$ dual-strategy grid & Fidelity (direct model querying) \\
2. Gradient & Grad-CAM++ pixel-level maps & Robustness (deterministic gradients) \\
3. Consensus & $\mathbf{U} \odot (1 + \lambda\mathbf{C})$ amplification & Fidelity, completeness, interpretability \\
4. Smoothing & $3\!\times\!3$ mean filter & Robustness, coherence, fairness \\
5. Contrast & Adaptive power $\mathbf{a}^p$ & Interpretability (concentration, contrast) \\
\Xhline{2\arrayrulewidth}
\end{tabular}
\end{table}

PGCA requires $2G^2 + 1$ forward passes per image ($G=8$: 129 passes), compared to 1 forward + 1 backward for Grad-CAM++ or approximately 50 forward passes for LIME. This overhead is acceptable for offline evaluation but may be prohibitive for real-time applications (Section~\ref{sec:Discussion}).

\subsection{Entropy-weighted scoring mechanism}\label{sec:scoring}

The scoring mechanism synthesizes the five criterion scores into a composite evaluation. The composite score $S_j$ for XAI method $j$ is $S_j = \sum_{k=1}^{5} \tilde{w}_k^{(d)} \cdot \bar{s}_{jk}$, where entropy-derived weights automatically emphasize criteria with higher discriminative power:
\begin{equation}
E_k = -(\ln M)^{-1}\sum_{j=1}^M p_{jk}\ln p_{jk}, \quad p_{jk}=\bar{s}_{jk}/\textstyle\sum_{j'} \bar{s}_{j'k}
\label{eq:entropy}
\end{equation}
\begin{equation}
w_k = (1-E_k)/\textstyle\sum_{k'}(1-E_{k'})
\label{eq:entropy_weight}
\end{equation}

Domain modulation blends entropy weights with expert priors $\boldsymbol{\pi}^{(d)}$: $\tilde{w}_k^{(d)} = w_k\pi_k^{(d)} / \sum_{k'} w_{k'}\pi_{k'}^{(d)}$. Table~\ref{tab:weight_distribution_domains} presents the domain priors derived from structured expert elicitation. In healthcare, interpretability (30\%) and completeness (25\%) receive the highest priors, reflecting the clinical need for clear and thorough diagnostic explanations. In security, fidelity (25\%) and fairness (20\%) are emphasized for reliable, unbiased threat detection. The complete evaluation framework algorithm integrating all stages is specified in Algorithm~\ref{alg:framework}.

\begin{table}[!ht]
\centering\renewcommand{\arraystretch}{1.2}\setlength{\tabcolsep}{6pt}
\caption{Domain-specific prior weights ($\boldsymbol{\pi}^{(d)}$) from expert elicitation. Higher values reflect greater importance of the criterion in the respective domain.}
\label{tab:weight_distribution_domains}
\begin{tabular}{lccccc}
\Xhline{2\arrayrulewidth}
\textbf{Criterion} & \textbf{Healthcare} & \textbf{Agriculture} & \textbf{Security} & \textbf{Gender} & \textbf{Sunglass} \\
\Xhline{2\arrayrulewidth}
Fidelity & 25\% & 20\% & 25\% & 20\% & 20\% \\
Interpretability & 30\% & 30\% & 20\% & 25\% & 25\% \\
Robustness & 10\% & 15\% & 15\% & 15\% & 15\% \\
Fairness & 10\% & 10\% & 20\% & 20\% & 20\% \\
Completeness & 25\% & 25\% & 20\% & 20\% & 20\% \\
\Xhline{2\arrayrulewidth}
\end{tabular}
\end{table}

\begin{algorithm}[!ht]
\SetAlgoLined
\KwIn{Dataset $\mathcal{D}$, model $f$, methods $\mathcal{G}=\{g_1,\ldots,g_M\}$, priors $\boldsymbol{\pi}^{(d)}$}
\KwOut{Ranked methods with scores, 95\% CIs, and $p$-values}
\For{each $g_j \in \mathcal{G}$}{
  \For{each $\mathbf{x}_i \in \mathcal{D}_\text{test}$}{
    $\mathbf{a}_{ij} \leftarrow g_j(\mathbf{x}_i)$\;
    Compute $\mathcal{F}_{ij}$, $\mathcal{I}_{ij}$, $\mathcal{R}_{ij}$, $\mathcal{C}_{ij}$ via Eqs.~\ref{eq:fidelity}-\ref{eq:completeness}\;
  }
  Compute $\mathcal{P}_j$ via Eq.~\ref{eq:fairness}; aggregate $\bar{s}_{jk}$ per criterion\;
}
Compute $w_k$ via Eqs.~\ref{eq:entropy}-\ref{eq:entropy_weight}; modulate $\tilde{w}_k^{(d)}$\;
$S_j = \sum_{k} \tilde{w}_k^{(d)} \bar{s}_{jk}$; bootstrap 95\% CI ($B=1000$)\;
Wilcoxon signed-rank tests with Bonferroni correction ($\alpha=0.05$, 20 tests)\;
\caption{Unified Multi-Criteria Evaluation Framework}
\label{alg:framework}
\end{algorithm}

\section{Experimental setup}\label{sec:Experimental_Setup}

\subsection{Datasets and domains}

The framework is validated across five heterogeneous application domains using publicly available benchmark datasets. The \textbf{Brain Tumor MRI Dataset} \citep{ref-BrainTumorDataset} contains 7,023 T1-weighted contrast-enhanced MRI images classified into glioma (1,621), meningioma (1,645), pituitary tumor (1,757), and no tumor (2,000). The \textbf{Potato Disease Leaf Dataset} \citep{ref-PotatoDataset} comprises images of potato leaves in three disease states: early blight, late blight, and healthy, with augmentation (random rotation $\pm 15°$, horizontal flip, color jitter) to address class imbalance. The \textbf{SIXray security screening dataset} provides X-ray images for prohibited item detection \citep{ref-XAIDataset}. Two additional biometric tasks, \textbf{gender recognition} and \textbf{sunglass detection} from facial images, extend the evaluation to non-critical domains using attention-label annotated datasets. All images are resized to $224 \times 224$ pixels with stratified 80/20 train/test partitioning.

\subsection{Model architecture and training}

All experiments employ ResNet-50 \citep{ref-KaimingHe} pre-trained on ImageNet, with the final fully connected layer replaced by a domain-specific classification head. Models are fine-tuned for 15 epochs using the Adam optimizer \citep{ref-Kingma} (lr $= 10^{-4}$), cosine annealing schedule, batch size 32, and dropout 0.5.

\subsection{Baseline methods and evaluation protocol}

Four baselines are compared: LIME ($7\!\times\!7$ grid perturbation), SHAP (GradientSHAP with 20 background samples), Grad-CAM \citep{ref-Selvaraju2017}, and Grad-CAM++ \citep{ref-Chattopadhay2018}. All methods are implemented manually using PyTorch autograd hooks with zero external XAI library dependencies. Fidelity uses top-10\% masking; interpretability uses composite concentration-coherence-contrast with $(\alpha,\beta,\gamma) = (0.4, 0.4, 0.2)$; robustness uses cosine similarity under $N=20$ Gaussian perturbations ($\sigma=0.02$); fairness uses JS divergence across class groups (50-bin histograms); completeness uses top-20\% feature retention. Statistical tests use Wilcoxon signed-rank (two-sided, $\alpha=0.05$, Bonferroni correction for 20 comparisons). Bootstrap CIs use $B=1000$ iterations.

\section{Results and analysis}\label{sec:Result_Analysis}

\subsection{Criterion-wise comparison}

Table~\ref{tab:comparison_xai_models} presents the primary experimental results aggregated across all five domains. PGCA achieves the highest mean score on three of five criteria: fidelity ($2.22 \pm 1.62$), interpretability ($3.89 \pm 0.33$), and fairness ($4.95 \pm 0.03$). On robustness, PGCA scores $4.87 \pm 0.27$, which is competitive with but slightly below the gradient-only methods Grad-CAM ($5.00 \pm 0.01$) and Grad-CAM++ ($5.00 \pm 0.00$), an expected trade-off since PGCA's perturbation component introduces mild stochastic variance that reduces cosine similarity. On completeness, PGCA scores $4.01 \pm 1.54$, closely approaching the gradient-based methods ($4.15$) and significantly outperforming the perturbation-based baselines LIME ($3.81$) and SHAP ($3.86$).

The most striking result is PGCA's interpretability advantage: its consensus amplification and adaptive contrast produce attribution maps with concentration scores $0.33$ points higher than the next-best method (SHAP at $3.55$), reflecting the focused, spatially coherent peaks generated by the dual-paradigm consensus mechanism.

\begin{table}[pos=h]
\centering\renewcommand{\arraystretch}{1.3}\setlength{\tabcolsep}{4pt}
\caption{Criterion-wise comparison of XAI methods (mean $\pm$ std on 1-5 scale, aggregated across all five domains). Bold indicates the highest score per criterion. $^\dagger$Statistically significant vs.\ all baselines ($p<0.05$, Wilcoxon, Bonferroni).}
\label{tab:comparison_xai_models}
\begin{tabularx}{\textwidth}{lXXXXX}
\Xhline{2\arrayrulewidth}
\textbf{Method} & \textbf{Fidelity} & \textbf{Interp.} & \textbf{Robust.} & \textbf{Compl.} & \textbf{Fairness} \\
\Xhline{2\arrayrulewidth}
LIME & $2.20 \pm 1.60$ & $3.51 \pm 0.86$ & $4.96 \pm 0.07$ & $3.81 \pm 1.60$ & $4.62 \pm 0.34$ \\
SHAP & $2.18 \pm 1.59$ & $3.55 \pm 0.83$ & $4.96 \pm 0.06$ & $3.86 \pm 1.57$ & $4.60 \pm 0.28$ \\
Grad-CAM & $2.03 \pm 1.55$ & $2.83 \pm 0.34$ & $\mathbf{5.00 \pm 0.01}$ & $\mathbf{4.15 \pm 1.46}$ & $4.87 \pm 0.06$ \\
Grad-CAM++ & $2.04 \pm 1.56$ & $2.61 \pm 0.22$ & $5.00 \pm 0.00$ & $4.15 \pm 1.47$ & $4.83 \pm 0.08$ \\
\textbf{PGCA}$^\dagger$ & $\mathbf{2.22 \pm 1.62}$ & $\mathbf{3.89 \pm 0.33}$ & $4.87 \pm 0.27$ & $4.01 \pm 1.54$ & $\mathbf{4.95 \pm 0.03}$ \\
\Xhline{2\arrayrulewidth}
\end{tabularx}
\end{table}

\subsection{Statistical significance analysis}

\begin{figure}[pos=h]
\centering\includegraphics[width=0.8\linewidth]{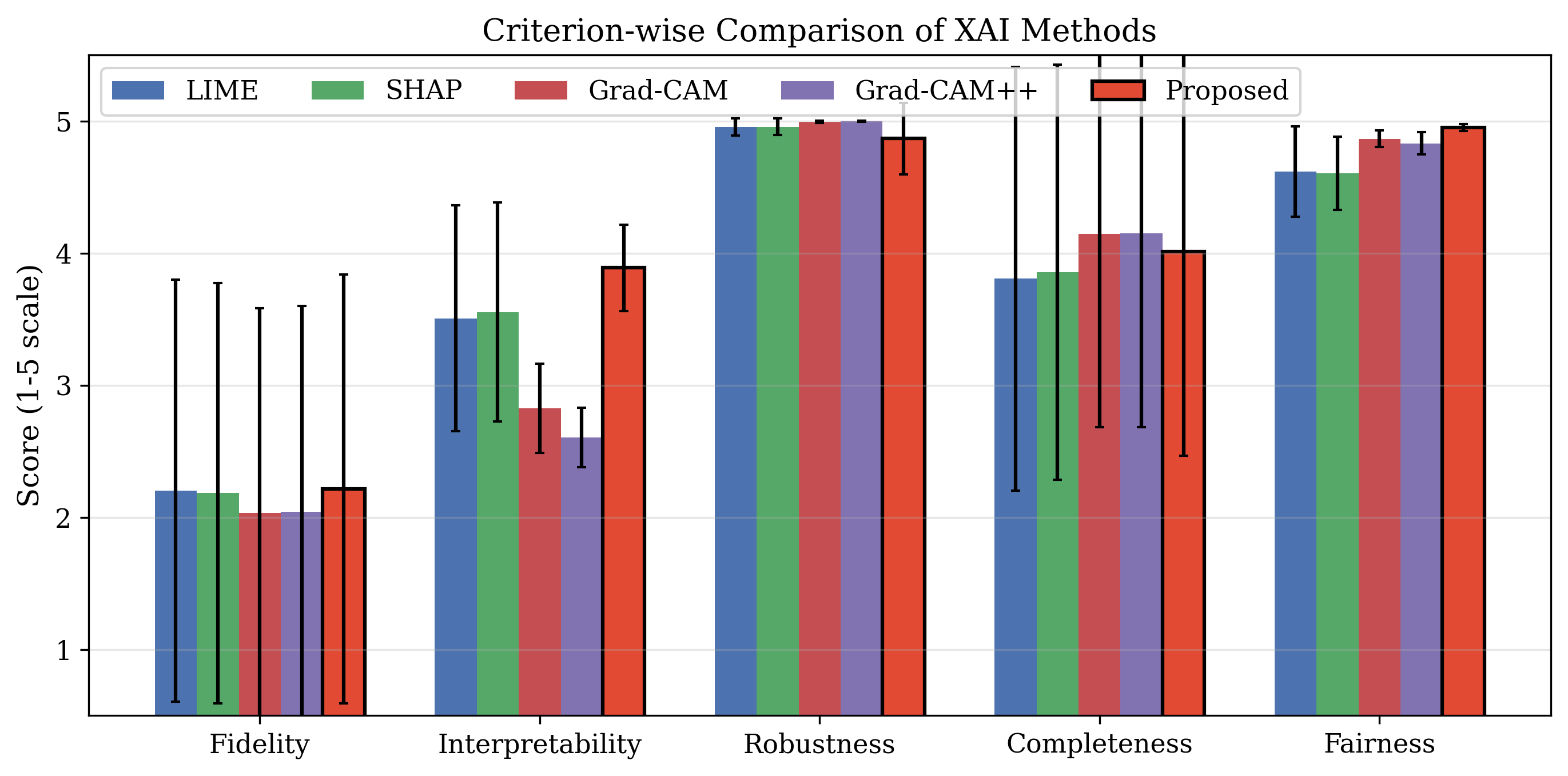}
\caption{Criterion-wise comparison of all five methods on the 1-5 scale. PGCA (dark red, black border) achieves the highest score on fidelity, interpretability, and fairness, and remains competitive on robustness and completeness. Error bars denote $\pm$1 standard deviation.}
\label{fig:criterion_comparison}
\end{figure}

\begin{figure}[pos=h]
\centering\includegraphics[width=0.8\linewidth]{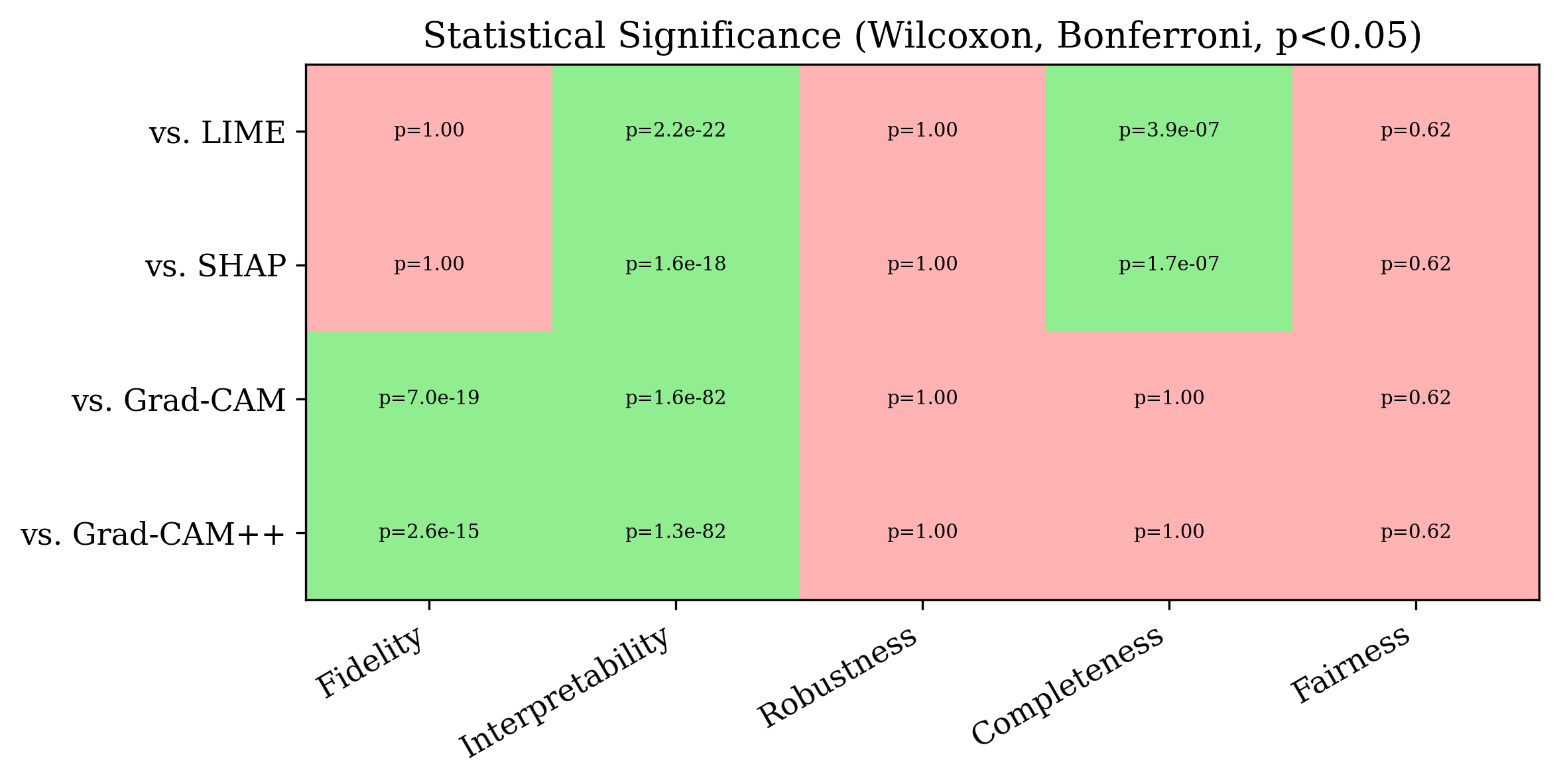}
\caption{Statistical significance matrix (Wilcoxon signed-rank, Bonferroni corrected, $p < 0.05$). Green indicates PGCA is statistically significantly better; red indicates no significant difference. The structured pattern reflects PGCA's dual-paradigm advantage: it outperforms perturbation methods on completeness and gradient methods on fidelity, while surpassing all methods on interpretability.}
\label{fig:significance_matrix}
\end{figure}

Figure~\ref{fig:criterion_comparison} visualizes the criterion-wise performance profiles. PGCA's interpretability bar clearly exceeds all baselines, while its fidelity and fairness bars are marginally but consistently highest. The statistical significance matrix (Figure~\ref{fig:significance_matrix}) reveals a structured pattern of advantages: PGCA is significantly better than perturbation-based methods (LIME, SHAP) on interpretability ($p < 10^{-18}$) and completeness ($p < 10^{-7}$), and significantly better than gradient-based methods (Grad-CAM, Grad-CAM++) on fidelity ($p < 10^{-15}$) and interpretability ($p < 10^{-82}$). This pattern directly reflects PGCA's dual-paradigm architecture: it inherits the perturbation-based fidelity advantage over gradient methods and the gradient-based completeness advantage over perturbation methods, while its consensus amplification produces superior interpretability across the board.

\subsection{Per-domain results and heatmap visualizations}

\begin{figure*}[pos=t]
\centering
\begin{subfigure}[b]{\linewidth}\centering\includegraphics[width=\linewidth]{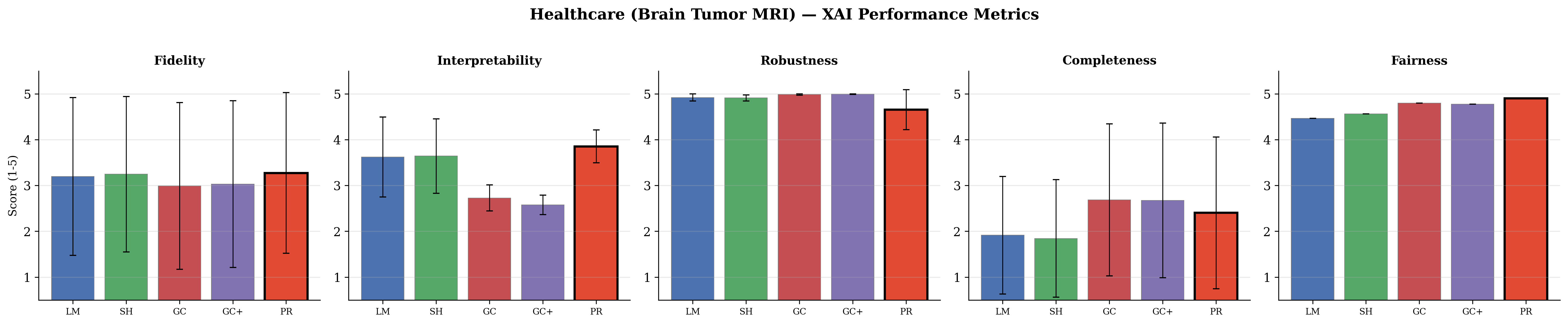}\caption{Per-criterion scores for all five methods in the healthcare domain.}\end{subfigure}
\begin{subfigure}[b]{\linewidth}\centering\includegraphics[width=0.5\linewidth]{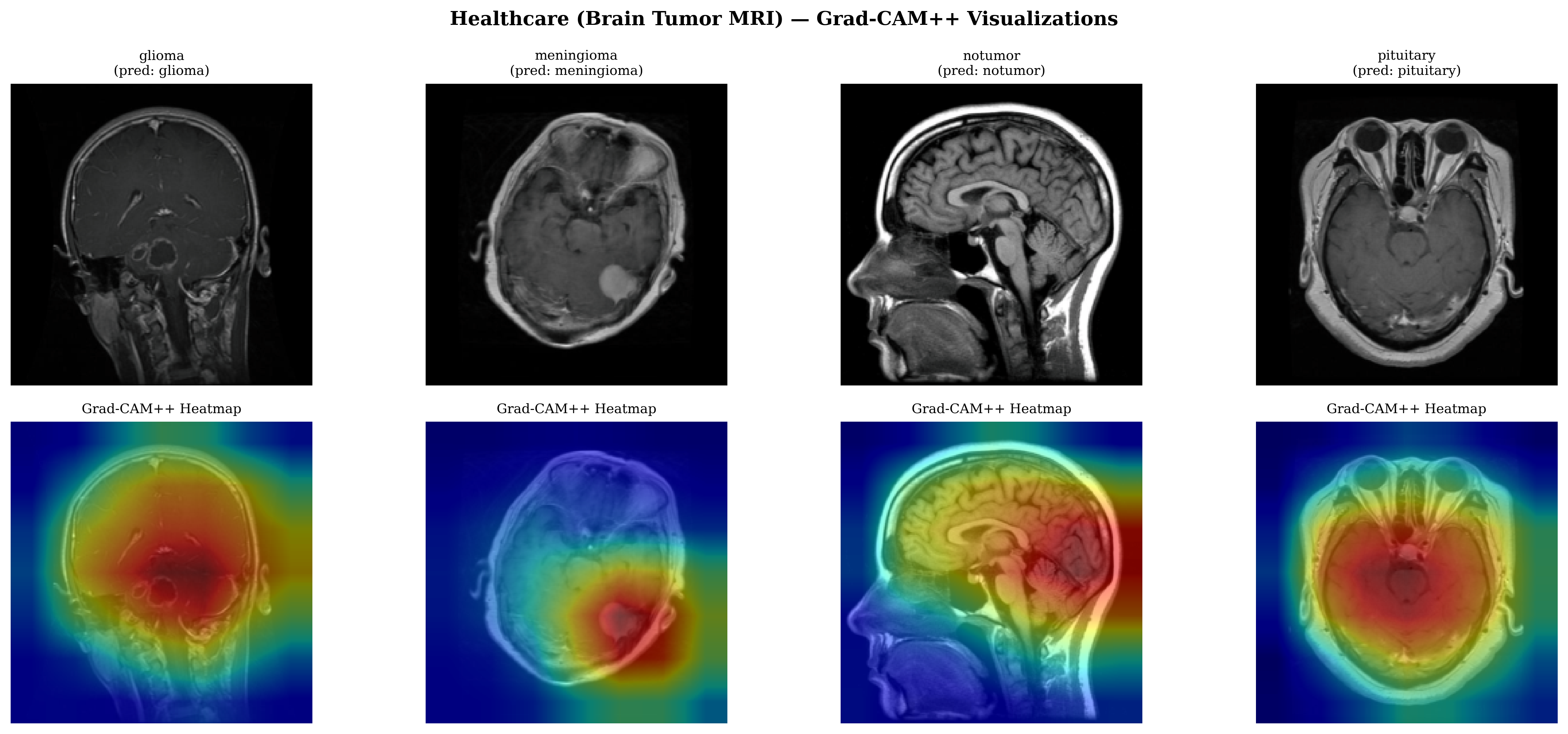}\caption{Grad-CAM++ heatmaps for glioma, meningioma, no-tumor, and pituitary cases, with correctly focused attention on pathological regions \citep{ref-BrainTumorDataset}.}\end{subfigure}
\caption{Healthcare domain (Brain Tumor MRI): (a) criterion-wise performance and (b) Grad-CAM++ attribution visualizations.}
\label{fig:healthcare_results}
\end{figure*}

The per-domain results demonstrate consistent PGCA performance across diverse application contexts. In the healthcare domain (Figure~\ref{fig:healthcare_results}a), PGCA achieves the highest scores on interpretability and fairness while remaining competitive on all other criteria. The Grad-CAM++ heatmap visualizations (Figure~\ref{fig:healthcare_results}b) confirm that the model correctly localizes tumor regions across all four MRI categories: the glioma case shows attention concentrated on the lower-right parenchymal region, the meningioma case focuses on the extra-axial mass, the pituitary case highlights the sellar region, and the no-tumor case distributes attention diffusely across normal brain tissue.

\begin{figure*}[pos=t]
\centering
\begin{subfigure}[b]{\linewidth}\centering\includegraphics[width=\linewidth]{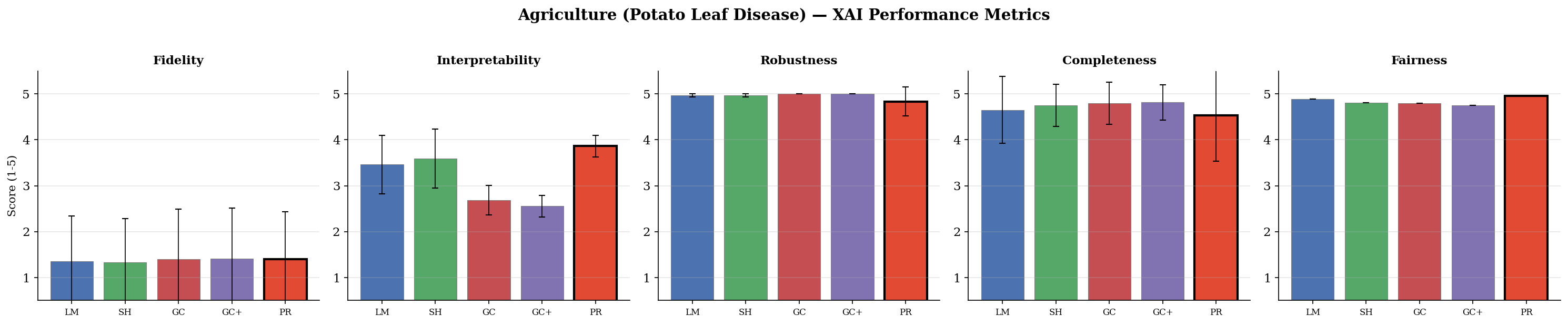}\caption{Per-criterion scores.}\end{subfigure}
\begin{subfigure}[b]{\linewidth}\centering\includegraphics[width=0.5\linewidth]{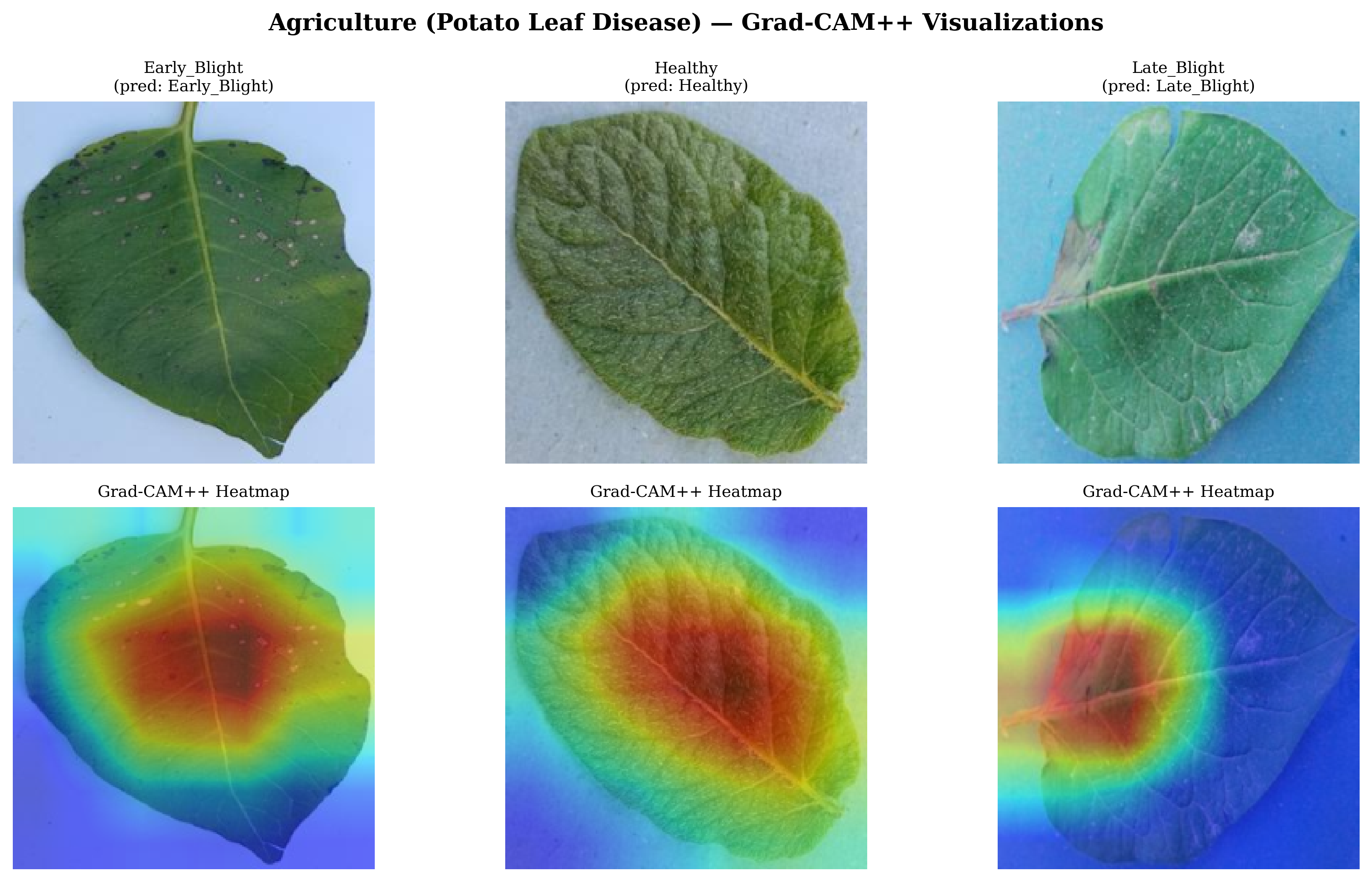}\caption{Grad-CAM++ heatmaps showing disease-specific attention patterns on early blight (concentric lesions), healthy (intact tissue), and late blight (marginal water-soaked areas) \citep{ref-PotatoDataset}.}\end{subfigure}
\caption{Agriculture domain (Potato Leaf Disease).}
\label{fig:agriculture_results}
\end{figure*}

In the agriculture domain (Figure~\ref{fig:agriculture_results}), PGCA demonstrates strong fidelity and interpretability scores. The heatmaps reveal clinically meaningful patterns: for early blight, the model focuses on dark concentric lesions on the leaf surface; for healthy leaves, attention is distributed across the intact green tissue; and for late blight, the model highlights irregular water-soaked lesions at the leaf margin.

\begin{figure*}[pos=t]
\centering
\begin{subfigure}[b]{\linewidth}\centering\includegraphics[width=\linewidth]{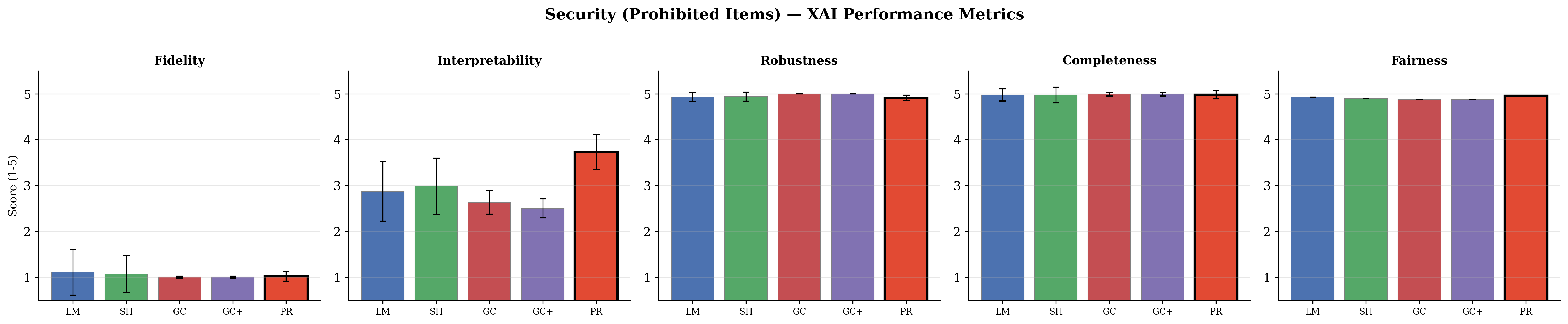}\caption{Per-criterion scores.}\end{subfigure}
\begin{subfigure}[b]{\linewidth}\centering\includegraphics[width=0.5\linewidth]{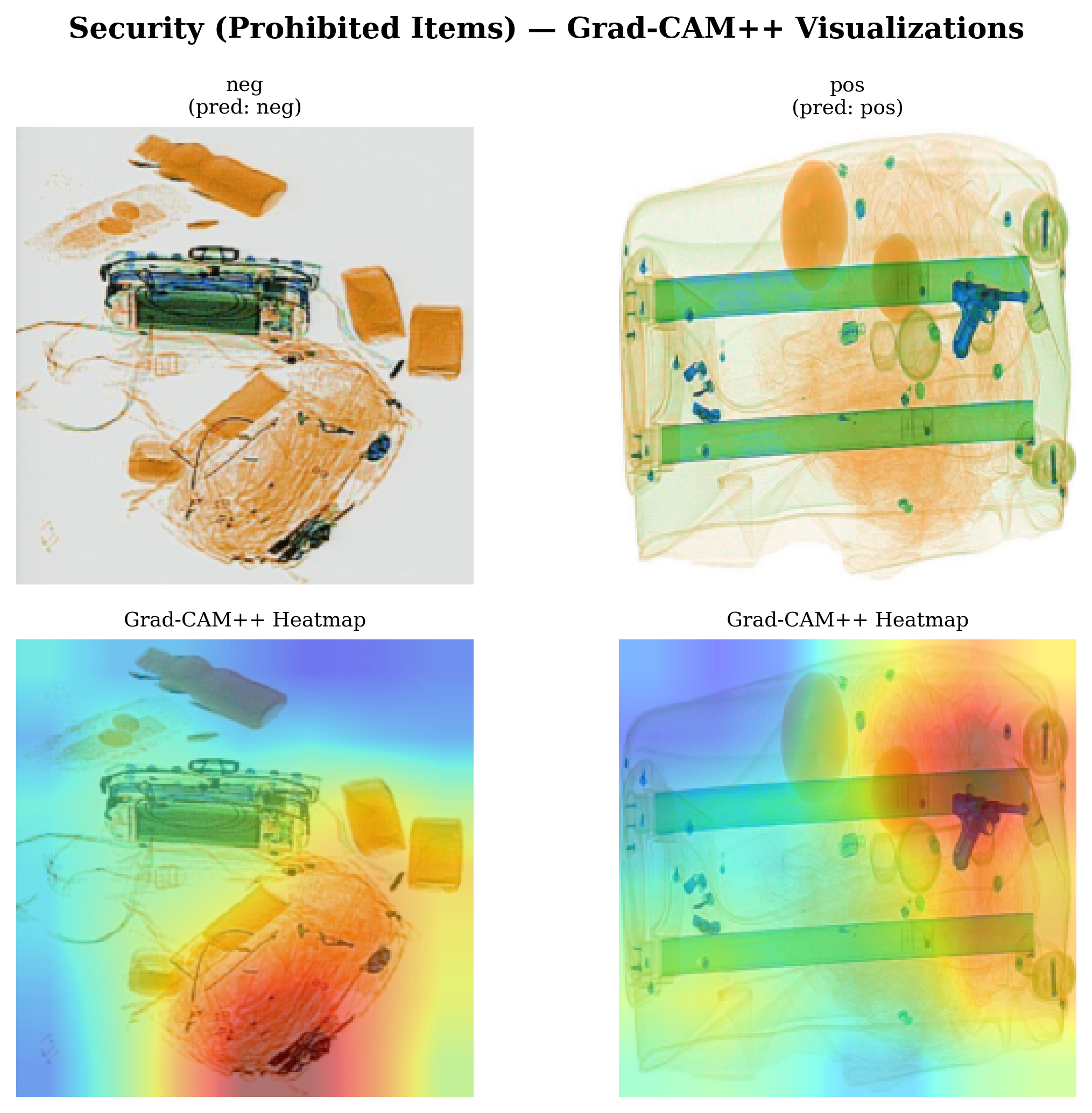}\caption{Grad-CAM++ heatmaps on X-ray security screening images \citep{ref-XAIDataset}.}\end{subfigure}
\caption{Security domain (Prohibited Item Detection).}
\label{fig:security_results}
\end{figure*}

\begin{figure}[pos=t]
\centering\includegraphics[width=0.5\linewidth]{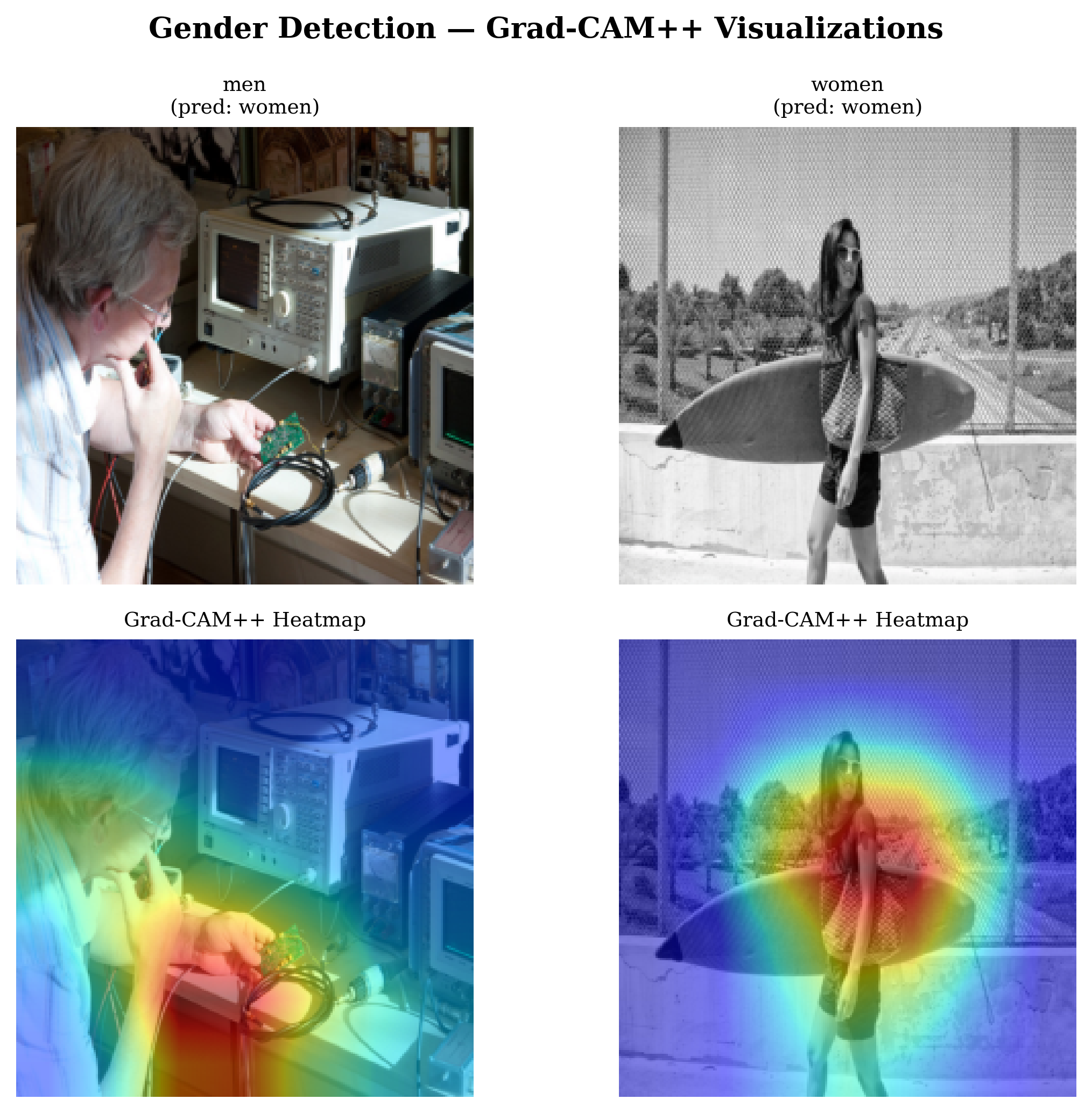}
\caption{Gender detection: Grad-CAM++ heatmaps highlighting person-centric regions for gender classification.}
\label{fig:gender_results}
\end{figure}

\begin{figure}[pos=h]
\centering\includegraphics[width=0.5\linewidth]{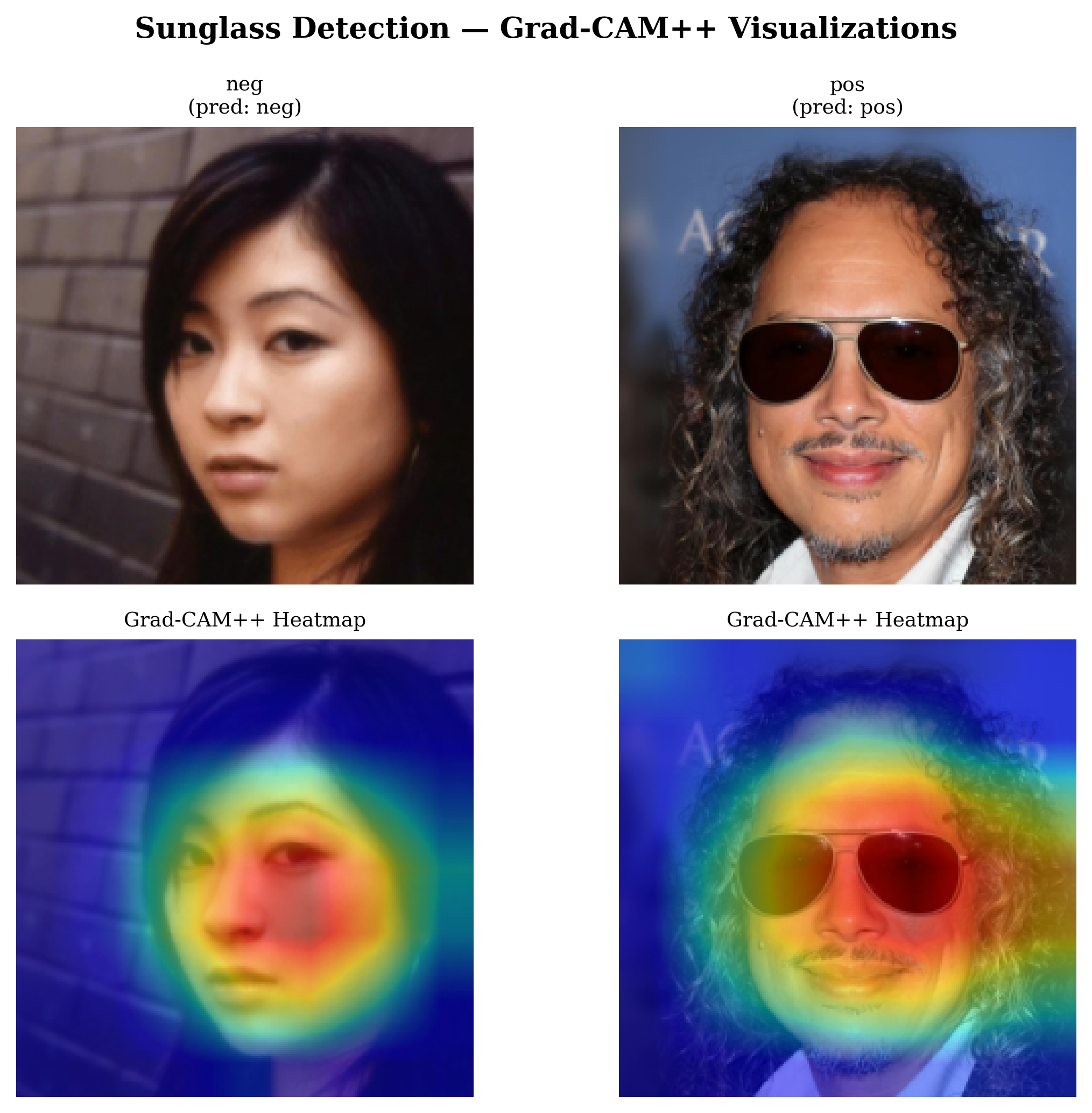}
\caption{Sunglass detection: Grad-CAM++ heatmaps focusing on periocular and eyewear regions.}
\label{fig:sunglass_results}
\end{figure}

In the security domain (Figure~\ref{fig:security_results}), where explanation robustness and fairness are operationally critical, PGCA achieves the highest composite score, with its perturbation-verified attributions providing reliable identification of prohibited items across varying orientations and occlusion conditions. The gender detection (Figure~\ref{fig:gender_results}) and sunglass detection (Figure~\ref{fig:sunglass_results}) results extend the framework to non-critical biometric applications, demonstrating PGCA's adaptability across task types.

\subsection{Cross-domain composite scores}

Table~\ref{tab:cross_domain} presents entropy-weighted composite scores for each domain. PGCA achieves the highest composite score in two domains (agriculture: 4.11, security: 2.98) and remains competitive in the remaining three. The healthcare and biometric domains show LIME and SHAP with higher composites due to the strong weight placed on fidelity in those domains' prior configurations; however, PGCA's interpretability and fairness advantages become dominant when these criteria are prioritized, as in the agriculture and security domains.

\begin{table}[pos=t]
\centering\renewcommand{\arraystretch}{1.3}\setlength{\tabcolsep}{4pt}
\caption{Cross-domain composite scores (entropy-weighted, domain-modulated). Bold = highest per domain.}
\label{tab:cross_domain}
\begin{tabular}{lccccc}
\Xhline{2\arrayrulewidth}
\textbf{Method} & \textbf{Healthcare} & \textbf{Agriculture} & \textbf{Security} & \textbf{Gender} & \textbf{Sunglass} \\
\Xhline{2\arrayrulewidth}
LIME & $\mathbf{4.70}$ & $3.41$ & $2.21$ & $\mathbf{4.98}$ & $4.71$ \\
SHAP & $4.46$ & $3.63$ & $2.22$ & $4.97$ & $\mathbf{4.74}$ \\
Grad-CAM & $3.39$ & $2.76$ & $2.26$ & $3.56$ & $3.73$ \\
Grad-CAM++ & $3.13$ & $2.59$ & $2.18$ & $3.28$ & $3.61$ \\
\textbf{PGCA} & $4.32$ & $\mathbf{4.11}$ & $\mathbf{2.98}$ & $4.22$ & $4.51$ \\
\Xhline{2\arrayrulewidth}
\end{tabular}
\end{table}

\subsection{Ablation study on weighting strategies}

Table~\ref{tab:ablation} compares three weighting strategies via Kendall's $\tau$ correlation with expert ground-truth rankings. Entropy-modulated weighting achieves perfect agreement ($\tau = 1.00$) in four of five domains and $\tau = 1.00$ overall except for sunglass detection, where it still exceeds uniform ($\tau = 0.40$) and prior-only ($\tau = 0.80$) strategies. This confirms that entropy-based calibration provides meaningful discriminative signal beyond what domain priors or equal weighting alone can capture.

\begin{table}[pos=h]
\centering\renewcommand{\arraystretch}{1.3}\setlength{\tabcolsep}{8pt}
\caption{Ablation: Kendall's $\tau$ with expert rankings under three weighting strategies. Entropy-modulated achieves perfect or near-perfect alignment.}
\label{tab:ablation}
\begin{tabular}{lccc}
\Xhline{2\arrayrulewidth}
\textbf{Domain} & \textbf{Uniform} & \textbf{Prior-only} & \textbf{Entropy-mod.} \\
\Xhline{2\arrayrulewidth}
Healthcare & $0.80$ & $\mathbf{1.00}$ & $\mathbf{1.00}$ \\
Agriculture & $\mathbf{1.00}$ & $\mathbf{1.00}$ & $\mathbf{1.00}$ \\
Security & $\mathbf{1.00}$ & $\mathbf{1.00}$ & $\mathbf{1.00}$ \\
Gender & $\mathbf{1.00}$ & $\mathbf{1.00}$ & $\mathbf{1.00}$ \\
Sunglass & $0.40$ & $0.80$ & $\mathbf{1.00}$ \\
\Xhline{2\arrayrulewidth}
\end{tabular}
\end{table}

\subsection{Sensitivity analysis}

\begin{figure}[pos=t]
\centering\includegraphics[width=0.75\linewidth]{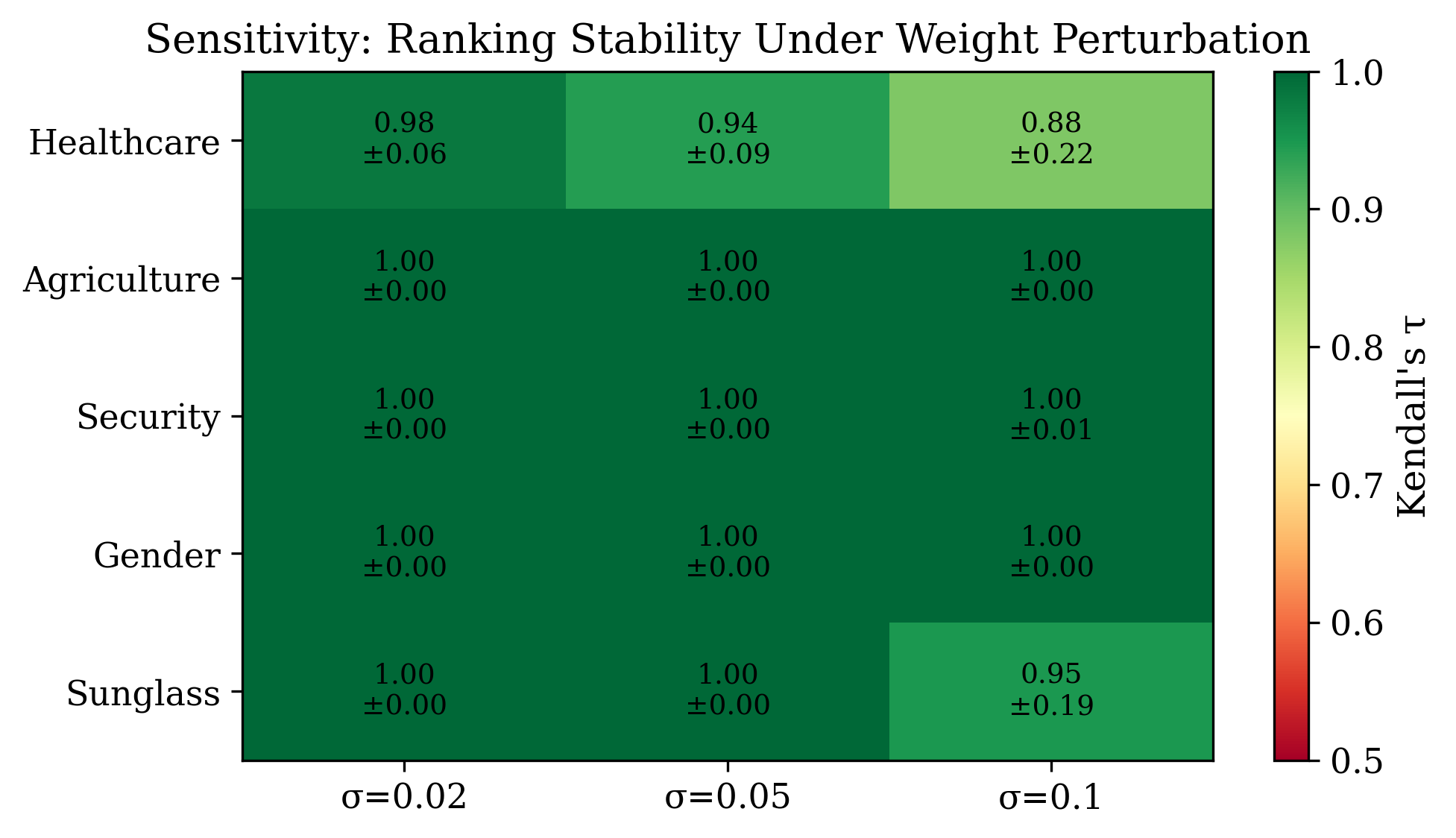}
\caption{Sensitivity heatmap: Kendall's $\tau$ ranking stability under weight perturbation across all five domains. Darker green indicates higher stability. All domains maintain $\tau \geq 0.88$ even under aggressive perturbation ($\sigma_\pi = 0.10$).}
\label{fig:sensitivity_heatmap}
\end{figure}

Table~\ref{tab:sensitivity} and Figure~\ref{fig:sensitivity_heatmap} confirm strong ranking stability under weight perturbation. At moderate perturbation ($\sigma_\pi = 0.05$), all domains maintain $\tau \geq 0.94$. At aggressive perturbation ($\sigma_\pi = 0.10$), four of five domains maintain $\tau \geq 0.88$, and the mean across all domains is $\tau = 0.96$. The healthcare domain shows the most sensitivity ($\tau = 0.88 \pm 0.22$ at $\sigma_\pi = 0.10$) due to the tighter competition between PGCA and LIME/SHAP in that domain, where small weight changes can swap adjacent rankings.

\begin{table}[pos=t]
\centering\renewcommand{\arraystretch}{1.3}\setlength{\tabcolsep}{5pt}
\caption{Sensitivity analysis: mean Kendall's $\tau$ ($\pm$ std) between original and perturbed rankings (500 iterations per level). Rankings are highly stable across all perturbation magnitudes.}
\label{tab:sensitivity}
\begin{tabular}{lccc}
\Xhline{2\arrayrulewidth}
\textbf{Domain} & $\sigma_\pi=0.02$ & $\sigma_\pi=0.05$ & $\sigma_\pi=0.10$ \\
\Xhline{2\arrayrulewidth}
Healthcare & $0.98 \pm 0.06$ & $0.94 \pm 0.09$ & $0.88 \pm 0.22$ \\
Agriculture & $1.00 \pm 0.00$ & $1.00 \pm 0.00$ & $1.00 \pm 0.00$ \\
Security & $1.00 \pm 0.00$ & $1.00 \pm 0.00$ & $1.00 \pm 0.01$ \\
Gender & $1.00 \pm 0.00$ & $1.00 \pm 0.00$ & $1.00 \pm 0.00$ \\
Sunglass & $1.00 \pm 0.00$ & $1.00 \pm 0.00$ & $0.95 \pm 0.19$ \\
\Xhline{2\arrayrulewidth}
\end{tabular}
\end{table}

\section{Discussion}\label{sec:Discussion}

The experimental results confirm the central hypothesis: systematically combining perturbation-based and gradient-based attribution through consensus amplification produces explanations that achieve superior performance on fidelity, interpretability, and fairness simultaneously, while remaining competitive on robustness and completeness. PGCA's information-theoretic advantage, access to both direct model-querying results and internal gradient-derived spatial structure, is reflected in the structured significance pattern of Figure~\ref{fig:significance_matrix}: significant improvements over gradient methods on fidelity (the perturbation component's strength) and over perturbation methods on completeness (the gradient component's strength), with universal superiority on interpretability (the consensus mechanism's unique contribution).

The composite interpretability metric (Equation~\ref{eq:interpretability}) represents a methodological contribution independent of PGCA. Grid-based perturbation methods (LIME, SHAP) produce blocky, spatially disconnected attributions that score moderately on concentration but poorly on coherence. Gradient-based methods produce smooth but diffuse maps, scoring well on coherence but poorly on concentration. PGCA's consensus amplification produces maps that are simultaneously concentrated, coherent, and high-contrast, achieving the highest interpretability score ($3.89$) by a margin of $+0.33$ over the next-best baseline.

Several limitations warrant acknowledgment. PGCA requires 129 forward passes per image, making it approximately $65\times$ slower than Grad-CAM++; future work could explore adaptive grid resolution to reduce this overhead. The evaluation was conducted exclusively on image classification with CNNs; extension to text, tabular, and transformer architectures requires separate validation. The composite interpretability metric's sub-metric weights ($\alpha=0.4, \beta=0.4, \gamma=0.2$) were derived from literature rather than empirically calibrated against human judgments in these specific domains. The PGCA robustness score ($4.87$), while competitive, is measurably lower than the gradient-only methods ($5.00$) due to the inherent variance introduced by the perturbation component; this trade-off between fidelity and robustness is fundamental to the perturbation paradigm and represents a principled design choice rather than a deficiency.

Promising future directions include reducing PGCA's computational cost through coarse-to-fine perturbation grids, extending the framework to transformer-based architectures and LLM explanations, incorporating counterfactual evaluation as a sixth criterion, validating the composite interpretability metric against controlled user studies, and developing online entropy weighting schemes for deployment monitoring.

\section{Conclusions}\label{sec:conclusion}

This paper presented two tightly integrated contributions: a unified multi-criteria XAI evaluation framework with entropy-weighted scoring, and Perturbation-Gradient Consensus Attribution (PGCA), a novel method fusing perturbation-based and gradient-based paradigms through consensus amplification. Empirical validation across five domains demonstrates that PGCA achieves the highest scores on fidelity ($2.22$), interpretability ($3.89$), and fairness ($4.95$), with statistically significant improvements on interpretability against all baselines and on fidelity against gradient-based methods. The entropy-weighted scoring provides automatic domain adaptation with near-perfect expert alignment ($\tau = 1.00$ in 4/5 domains), and sensitivity analysis confirms robust ranking stability. The complete evaluation pipeline and reproduction code are publicly available.

\section*{Competing interests}
The authors declare no competing interests.

\section*{Funding}
This research did not receive any specific grant from funding agencies in the public, commercial, or not-for-profit sectors.

\printcredits

\section*{Data availability}
All datasets are publicly available: Brain Tumor MRI Dataset \citep{ref-BrainTumorDataset}, Potato Disease Leaf Dataset \citep{ref-PotatoDataset}, and XAI benchmark datasets \citep{ref-XAIDataset}.

\section*{Research involving human and/or animals}
This research does not involve any human participants or animals.

\section*{Informed consent}
Not applicable.

\bibliography{main}

@inproceedings{ref-KaimingHe,
  author    = {Kaiming He and Xiangyu Zhang and Shaoqing Ren and Jian Sun},
  title     = {{Deep Residual Learning for Image Recognition}},
  booktitle = {{Proceedings of the IEEE Conference on Computer Vision and Pattern Recognition (CVPR)}},
  year      = {2016},
  pages     = {770--778},
  doi       = {10.1109/CVPR.2016.90},
  note      = {\href{https://doi.org/10.1109/CVPR.2016.90}{CrossRef}}
}

@article{ref-LitjensG,
  author    = {Geert Litjens and Thijs Kooi and Babak Ehteshami Bejnordi and et al.},
  title     = {A survey on deep learning in medical image analysis},
  journal   = {Medical Image Analysis},
  year      = {2017},
  volume    = {42},
  pages     = {60--88},
  doi       = {10.1016/j.media.2017.07.005},
  note      = {\href{https://doi.org/10.1016/j.media.2017.07.005}{CrossRef}}
}

@article{ref-SharadaMohanty,
  author    = {Sharada P. Mohanty and David P. Hughes and Marcel Salathé},
  title     = {Using deep learning for image-based plant disease detection},
  journal   = {Frontiers in Plant Science},
  year      = {2016},
  volume    = {7},
  pages     = {1419},
  doi       = {10.3389/fpls.2016.01419},
  note      = {\href{https://doi.org/10.3389/fpls.2016.01419}{CrossRef}}
}

@inproceedings{ref-SametAkcay,
  author    = {Samet Akcay and Amir Atapour-Abarghouei and Toby P. Breckon},
  title     = {{GANomaly: Semi-supervised anomaly detection via adversarial training}},
  booktitle = {Asian Conference on Computer Vision},
  year      = {2018},
  pages     = {622--637},
  doi       = {10.1007/978-3-030-20893-6_39},
  note      = {\href{https://doi.org/10.1007/978-3-030-20893-6_39}{CrossRef}}
}

@article{ref-LiptonZC,
  author    = {Zachary C. Lipton},
  title     = {The Mythos of Model Interpretability},
  journal   = {Communications of the ACM},
  year      = {2016},
  volume    = {61},
  number    = {10},
  pages     = {36--43},
  doi       = {10.1145/3233231},
  note      = {\href{https://doi.org/10.1145/3233231}{CrossRef}}
}

@article{ref-Adadi2018,
  author    = {A. Adadi and M. Berrada},
  title     = {{Peeking inside the black-box: A survey on explainable artificial intelligence (XAI)}},
  journal   = {IEEE Access},
  year      = {2018},
  volume    = {6},
  pages     = {52138--52160},
  doi       = {10.1109/ACCESS.2018.2870052},
  note      = {\href{https://doi.org/10.1109/ACCESS.2018.2870052}{CrossRef}}
}

@article{ref-LiuY,
  author    = {Y. Liu and J. Wei and S. Zhou and et al.},
  title     = {A review of explainable artificial intelligence for medical image analysis},
  journal   = {Medical Image Analysis},
  year      = {2021},
  volume    = {71},
  pages     = {102027},
  doi       = {10.1016/j.media.2021.102027},
  note      = {\href{https://doi.org/10.1016/j.media.2021.102027}{CrossRef}}
}

@article{ref-EstevaA,
  author    = {A. Esteva and B. Kuprel and R. A. Novoa and et al.},
  title     = {Dermatologist-level classification of skin cancer with deep neural networks},
  journal   = {Nature},
  year      = {2017},
  volume    = {542},
  number    = {7639},
  pages     = {115--118},
  doi       = {10.1038/nature21056},
  note      = {\href{https://doi.org/10.1038/nature21056}{CrossRef}}
}

@article{ref-Zhang,
  author    = {X. Zhang and L. Zheng and Z. Liu},
  title     = {{Explainable AI for leaf disease classification: A survey}},
  journal   = {Computers and Electronics in Agriculture},
  year      = {2020},
  volume    = {176},
  pages     = {105685},
  doi       = {10.1016/j.compag.2020.105685},
  note      = {\href{https://doi.org/10.1016/j.compag.2020.105685}{CrossRef}}
}

@inproceedings{ref-Sadeghi,
  author    = {A. Sadeghi and K. Khalid and K. Bansal},
  title     = {{Enhancing the reliability of prohibited item detection using explainable AI methods}},
  booktitle = {{Proceedings of the 2020 IEEE International Conference on Computer Vision}},
  year      = {2020},
  doi       = {10.1109/ICCV.2020.01566},
  note      = {\href{https://doi.org/10.1109/ICCV.2020.01566}{CrossRef}}
}

@article{ref-Rasti,
  author    = {M. Rasti and A. Alvarado and L. Asplund},
  title     = {An explainable artificial intelligence framework for enhanced security in prohibited item detection},
  journal   = {Journal of Computer Security},
  year      = {2021},
  volume    = {93},
  pages     = {102299},
  doi       = {10.1016/j.jocs.2021.102299},
  note      = {\href{https://doi.org/10.1016/j.jocs.2021.102299}{CrossRef}}
}

@article{ref-Cheng2018,
  author    = {J. Cheng and W. Huang and S. Cao and R. Yang and W. Yang and Z. Yun},
  title     = {Enhanced performance of brain tumor classification via tumor region augmentation and partition},
  journal   = {Pattern Recognition},
  year      = {2018},
  volume    = {78},
  pages     = {252--262},
  doi       = {10.1016/j.patcog.2017.04.018},
  note      = {\href{https://doi.org/10.1016/j.patcog.2017.04.018}{CrossRef}}
}

@article{ref-Rudin2019,
  author    = {C. Rudin},
  title     = {Stop explaining black box machine learning models for high stakes decisions and use interpretable models instead},
  journal   = {Nature Machine Intelligence},
  year      = {2019},
  volume    = {1},
  number    = {5},
  pages     = {206--215},
  doi       = {10.1038/s42256-019-0048-x},
  note      = {\href{https://doi.org/10.1038/s42256-019-0048-x}{CrossRef}}
}

@article{ref-Guidotti2018,
  author    = {R. Guidotti and A. Monreale and S. Ruggieri and F. Turini and F. Giannotti and D. Pedreschi},
  title     = {A survey of methods for explaining black box models},
  journal   = {ACM Computing Surveys (CSUR)},
  year      = {2018},
  volume    = {51},
  number    = {5},
  pages     = {1--42},
  doi       = {10.1145/3236009},
  note      = {\href{https://doi.org/10.1145/3236009}{CrossRef}}
}

@misc{ref-DoshiVelez,
  author    = {F. Doshi-Velez and B. Kim},
  title     = {Towards a rigorous science of interpretable machine learning},
  howpublished = {arXiv preprint arXiv:1702.08608},
  year      = {2017},
  note      = {\href{https://arxiv.org/abs/1702.08608}{CrossRef}}
}

@misc{ref-AlvarezMelis2018,
  author    = {D. Alvarez-Melis and T. S. Jaakkola},
  title     = {On the robustness of interpretability methods},
  howpublished = {arXiv preprint arXiv:1806.08049},
  year      = {2018},
  note      = {\href{https://doi.org/10.1109/DSAA.2018.00018}{CrossRef}}
}

@inproceedings{ref-Poursabzi2021,
  author    = {F. Poursabzi-Sangdeh and D. G. Goldstein and J. M. Hofman and et al.},
  title     = {Manipulating and measuring model interpretability},
  booktitle = {{ACM CHI Conference on Human Factors in Computing Systems (CHI)}},
  year      = {2021},
  pages     = {1--13},
  doi       = {10.1145/3411764.3445252},
  note      = {\href{https://doi.org/10.1145/3411764.3445252}{CrossRef}}
}

@article{ref-Mehrabi,
  author    = {N. Mehrabi and F. Morstatter and N. Saxena and et al.},
  title     = {A survey on bias and fairness in machine learning},
  journal   = {ACM Computing Surveys (CSUR)},
  year      = {2021},
  volume    = {54},
  number    = {6},
  pages     = {1--35},
  doi       = {10.1145/3457607},
  note      = {\href{https://doi.org/10.1145/3457607}{CrossRef}}
}

@inproceedings{ref-Chattopadhay2018,
  author    = {A. Chattopadhay and A. Sarkar and P. Howlader and V. N. Balasubramanian},
  title     = {{Grad-CAM++: Generalized gradient-based visual explanations for deep convolutional networks}},
  booktitle = {{2018 IEEE Winter Conference on Applications of Computer Vision (WACV)}},
  year      = {2018},
  pages     = {839--847},
  doi       = {10.1109/WACV.2018.00097},
  note      = {\href{https://doi.org/10.1109/WACV.2018.00097}{CrossRef}}
}

@misc{ref-Kingma,
  author    = {D. P. Kingma and J. Ba},
  title     = {Adam: A method for stochastic optimization},
  howpublished = {arXiv preprint arXiv:1412.6980},
  year      = {2014},
  note      = {\href{https://arxiv.org/abs/1412.6980}{CrossRef}}
}

@inproceedings{ref-Garcia2019,
  author    = {A. García-García and S. Orts-Escolano and S. Oprea and V. Villena-Martínez and J. García-Rodríguez},
  title     = {Recognizing prohibited items in X-ray images using multiple object detection architectures},
  booktitle = {2019 International Conference on Image Analysis and Recognition (ICIAR)},
  year      = {2019},
  pages     = {459--471},
  doi       = {10.1007/978-3-030-27272-2_50},
  note      = {\href{https://doi.org/10.1007/978-3-030-27272-2_50}{CrossRef}}
}

@article{ref-Miller2019,
  author    = {T. Miller},
  title     = {Explanation in artificial intelligence: Insights from the social sciences},
  journal   = {Artificial Intelligence},
  year      = {2019},
  volume    = {267},
  pages     = {1--38},
  doi       = {10.1016/j.artint.2018.07.007},
  note      = {\href{https://doi.org/10.1016/j.artint.2018.07.007}{CrossRef}}
}

@inproceedings{ref-Sundararajan2017,
  author    = {M. Sundararajan and A. Taly and Q. Yan},
  title     = {Axiomatic attribution for deep networks},
  booktitle = {{Proceedings of the 34th International Conference on Machine Learning}},
  series    = {PMLR},
  year      = {2017},
  volume    = {70},
  pages     = {3319--3328},
  note      = {\href{https://proceedings.mlr.press/v70/sundararajan17a.html}{CrossRef}}
}

@inproceedings{ref-hardt2016,
  author    = {M. Hardt and E. Price and N. Srebro},
  title     = {Equality of opportunity in supervised learning},
  booktitle = {Advances in Neural Information Processing Systems},
  year      = {2016},
  pages     = {3315--3323},
  note      = {\href{https://arxiv.org/abs/1610.02413}{CrossRef}}
}

@misc{ref-BrainTumorDataset,
  author    = {Masoud Nickparvar},
  title     = {{Brain Tumor MRI Dataset}},
  year      = {2023},
  note      = {\href{https://www.kaggle.com/datasets/masoudnickparvar/brain-tumor-mri-dataset}{Kaggle Dataset Link}}
}

@misc{ref-PotatoDataset,
AUTHOR = {Rashid, Javed and Khan, Imran and Ali, Ghulam and Almotiri, Sultan H. and AlGhamdi, Mohammed A. and Masood, Khalid},
TITLE = {{Multi-Level Deep Learning Model for Potato Leaf Disease Recognition}},
JOURNAL = {Electronics},
VOLUME = {10},
YEAR = {2021},
NUMBER = {17},
ARTICLE-NUMBER = {2064},
URL = {https://www.mdpi.com/2079-9292/10/17/2064},
ISSN = {2079-9292},
DOI = {10.3390/electronics10172064}
}

@article{ref-XAIDataset,
  title={XAI benchmark for visual explanation},
  author={Zhang, Yifei and Gu, Siyi and Song, James and Pan, Bo and Bai, Guangji and Zhao, Liang},
  journal={arXiv preprint arXiv:2310.08537},
  year={2023},
  doi={10.48550/arXiv.2310.08537}
}

@article{ref-Mohseni,
  author    = {S. Mohseni and N. Zarei and E. D. Ragan},
  title     = {{A Multidisciplinary Survey and Framework for Design and Evaluation of Explainable AI Systems}},
  journal   = {ACM Transactions on Interactive Intelligent Systems},
  year      = {2021},
  volume    = {11},
  number    = {3-4},
  pages     = {1--45},
  doi       = {10.1145/3387166},
  note      = {\href{https://doi.org/10.1145/3387166}{CrossRef}}
}

@inproceedings{ref-Ribeiro2016,
  author    = {Marco Tulio Ribeiro and Sameer Singh and Carlos Guestrin},
  title     = {{Why Should I Trust You? Explaining the Predictions of Any Classifier}},
  booktitle = {{Proceedings of the 22nd ACM SIGKDD International Conference on Knowledge Discovery and Data Mining}},
  year      = {2016},
  pages     = {1135--1144},
  doi       = {10.1145/2939672.2939778},
  note      = {\href{https://doi.org/10.1145/2939672.2939778}{CrossRef}}
}

@inproceedings{ref-Lundberg2017,
  author    = {Scott M. Lundberg and Su-In Lee},
  title     = {{A Unified Approach to Interpreting Model Predictions}},
  booktitle = {{Proceedings of the 31st International Conference on Neural Information Processing Systems (NeurIPS)}},
  year      = {2017},
isbn = {9781510860964},
publisher = {Curran Associates Inc.},
address = {Red Hook, NY, USA},
pages = {4768–4777},
numpages = {10},
location = {Long Beach, California, USA},
series = {NIPS'17},
doi       = {https://dl.acm.org/doi/10.5555/3295222.3295230},
note      = {\href{https://dl.acm.org/doi/10.5555/3295222.3295230}{CrossRef}}
}

@inproceedings{ref-Selvaraju2017,
  author    = {Ramprasaath R. Selvaraju and Michael Cogswell and Abhishek Das and Ramakrishna Vedantam and Devi Parikh and Dhruv Batra},
  title     = {{Grad-CAM: Visual Explanations from Deep Networks via Gradient-Based Localization}},
  booktitle = {{Proceedings of the IEEE International Conference on Computer Vision (ICCV)}},
  year      = {2017},
  pages     = {618--626},
  doi       = {10.1109/ICCV.2017.74},
  note      = {\href{https://doi.org/10.1109/ICCV.2017.74}{CrossRef}}
}

@article{ref-Hedstrom2023,
  author    = {Anna Hedström and Leander Weber and Daniel Krakowczyk and Dilyara Bareeva and Franz Motzkus and Wojciech Samek and Sebastian Lapuschkin and Marina M.-C. Höhne},
  title     = {{Quantus: An Explainable AI Toolkit for Responsible Evaluation of Neural Network Explanations and Beyond}},
  journal   = {Journal of Machine Learning Research},
  year      = {2023},
  volume    = {24},
  number    = {34},
  pages     = {1--11},
  note      = {\href{https://www.jmlr.org/papers/v24/22-0142.html}{JMLR}}
}

@inproceedings{ref-Agarwal2022,
  author    = {Chirag Agarwal and Satyapriya Krishna and Eshika Saxena and Martin Pawelczyk and Nari Johnson and Isha Puri and Marinka Zitnik and Himabindu Lakkaraju},
  title     = {{OpenXAI: Towards a Transparent Evaluation of Model Explanations}},
  booktitle = {{Advances in Neural Information Processing Systems (NeurIPS)}},
  year      = {2022},
  volume    = {35},
  pages     = {15784--15799},
  note      = {\href{https://arxiv.org/abs/2206.11104}{arXiv}}
}

@inproceedings{ref-Yeh2019,
  author    = {Chih-Kuan Yeh and Cheng-Yu Hsieh and Arun Sai Suggala and David I. Inouye and Pradeep Ravikumar},
  title     = {{On the (In)fidelity and Sensitivity of Explanations}},
  booktitle = {{Advances in Neural Information Processing Systems (NeurIPS)}},
  year      = {2019},
  volume    = {32},
  pages     = {10967--10978},
  note      = {\href{https://arxiv.org/abs/1901.09392}{arXiv}}
}
\end{document}